\documentclass[letterpaper, 10 pt, conference]{ieeeconf}  % Comment this line out if you need a4paper
\usepackage{xcolor}
\usepackage{subcaption}
\usepackage{tikz}
\usepackage{booktabs}
\usepackage{tabularx}
\usepackage{graphicx} % for \rotatebox
\usetikzlibrary{calc,arrows.meta}
\definecolor{cNear}{HTML}{42A5F5}
\definecolor{cVeryNear}{HTML}{1565C0}
\definecolor{cInFront}{HTML}{2E7D32}
\definecolor{cBehind}{HTML}{8E24AA}
\definecolor{cLeft}{HTML}{00ACC1}
\definecolor{cRight}{HTML}{EF6C00}
\definecolor{cFrontLeft}{HTML}{7CB342}
\definecolor{cFrontRight}{HTML}{F9A825}
\definecolor{cRearLeft}{HTML}{6D4C41}
\definecolor{cRearRight}{HTML}{D81B60}
\definecolor{cOverlap}{HTML}{3949AB}
\definecolor{cTouching}{HTML}{00897B}

\definecolor{cFollows}{HTML}{1565C0}
\definecolor{cOvertakes}{HTML}{E53935}
\definecolor{cMerge}{HTML}{FB8C00}
\definecolor{cCross}{HTML}{00897B}
\definecolor{cLaneChange}{HTML}{6A1B9A}

\definecolor{cTLRelevant}{RGB}{216,27,96}
\definecolor{cTLControls}{RGB}{0,96,100}
\definecolor{signalRed}{RGB}{211,47,47}
\definecolor{signalGreen}{RGB}{46,125,50}

\newcommand{\halfRGcircle}{%
\begin{tikzpicture}[baseline=-0.5ex]
    \clip (0,0) circle (0.6ex);
    \fill[signalRed] (-0.7ex,-0.7ex) rectangle (0,0.7ex);
    \fill[signalGreen] (0,-0.7ex) rectangle (0.7ex,0.7ex);
    \draw[white, line width=0.3pt] (0,0) circle (0.6ex);
\end{tikzpicture}%
}
\definecolor{cInIntersection}{RGB}{117,117,117}

\newcommand{\legendline}[2]{%
    \textcolor{#1}{\rule[0.35ex]{1.5em}{1.2pt}}\,
    #2%
}

\newcommand{\pred}[1]{\texttt{#1}}
\newcommand{\defiff}{\mathrel{:\!\Leftrightarrow}}

\IEEEoverridecommandlockouts                              % This command is only needed if 
\usepackage{amsmath} % assumes amsmath package installed
\makeatletter
\let\labelindent\relax
\makeatother
\usepackage{enumitem}
\usepackage{cite}

\makeatletter
\let\NAT@parse\undefined
\makeatother

\usepackage[hidelinks]{hyperref}

\title{\LARGE \bf
Grounding Vision--Language Models in Driving Semantics:

A Multi-Dataset Predicate Framework for Explainable Reasoning}

\author{Mohamed Chouai$^{1}$, Fazli Faruk Okumus$^{1}$, and Stefan Kugele$^{1}$%
\thanks{$^{1}$ AImotion Bavaria,
        Technische Hochschule Ingolstadt, 85049 Ingolstadt, Germany.
        {\tt\small mohamed.chouai@thi.de}}%}%
}
\begin{document}

\maketitle
\thispagestyle{empty}
\pagestyle{empty}

%%%%%%%%%%%%%%%%%%%%%%%%%%%%%%%%%%%%%%%%%%%%%%%%%%%%%%%%%%%%%%%%%%%%%%%%%%%%%%%%
\begin{abstract}

Vision--language models are increasingly used for driving-scene understanding, yet the semantic relations expressed in their outputs are often difficult to verify against the underlying traffic situation. This paper introduces a deterministic multi-dataset predicate framework that derives driving-scene semantics from measurable geometric, kinematic, temporal, map, and traffic-control evidence. Dataset-specific interfaces are used only to recover the required scene information, while predicate definitions remain unchanged across nuPlan and nuScenes and are materialised in a common Predicate Knowledge Graph. Quantitative semantic validation against manually annotated predicate relations on 200 scenarios from each dataset yields macro $F_1$ scores of 0.94 on nuPlan and 0.93 on nuScenes, with an average cross-dataset difference of 0.02 across the shared predicates. The Predicate KG is further evaluated using a frozen LLaVA-OneVision-7B model on the nine NuPlanQA subtasks. Predicate grounding achieves the highest accuracy among the evaluated visual-input conditions in seven of nine NuPlanQA subtasks, including Traffic Light (53.2\% to 71.5\%), Situation Assessment (76.2\% to 86.1\%), and Action Recommendation (82.9\% to 89.0\%). Weather/Lighting remains essentially unchanged (89.4\% vs.\ 88.8\%), consistent with the absence of corresponding predicates, while Predicate KG only input outperforms metadata-only input in eight of nine subtasks. The results show that deterministic predicates provide a consistent and traceable semantic representation and, under oracle grounding, can reduce visual dependence for reasoning tasks covered by the predicate vocabulary.

\end{abstract}

\section{Introduction}
\label{sec:introduction}

Recent progress in vision--language models (VLMs) has expanded the scope of autonomous-driving perception from object detection and scene recognition toward question answering, scene description, and natural-language reasoning~\cite{xu2024drivegpt4,sima2024drivelm,marcu2024lingoqa}. These developments are particularly relevant for explainable driving systems, where understanding a traffic scene requires more than identifying individual objects. Relations such as whether a vehicle is \textit{inFrontOf} another, whether it \textit{follows} the ego vehicle, or whether a pedestrian \textit{crossesInFrontOf} it depend on geometric configuration, motion, map context, and temporal evolution.

Semantic relations between traffic participants and road elements are important for situation understanding, behaviour prediction, risk assessment, and decision-making in autonomous driving. Yet the semantic statements produced by VLMs are difficult to verify directly against the underlying scene. A relation expressed in natural language may be plausible while still being inconsistent with the actual geometry or dynamics of the traffic situation. This is a relevant limitation in autonomous driving, where many semantic relations have a measurable physical basis and can therefore be defined explicitly. Structured representations such as ontologies~\cite{buechel2017ontology}, knowledge graphs~\cite{halilaj2021knowledge,mlodzian2023nusceneskg}, and traffic scene graphs~\cite{zipfl2022relation} provide mechanisms for expressing relations between traffic participants and road elements. However, such representations are commonly designed for particular datasets or downstream tasks and do not necessarily provide explicit operational definitions for the relations they encode.

A difficulty is the heterogeneity of autonomous-driving datasets. Object taxonomies, map representations, trajectory information, coordinate systems, and annotation formats differ considerably across datasets~\cite{halilaj2023integration}. As a result, semantically equivalent relations may be represented differently or may need to be reconstructed from different sources of information. This limits the use of dataset-specific semantics for systematic reasoning and evaluation across datasets.

\begin{figure*}[t]
    \centering
    \includegraphics[width=0.9\textwidth]{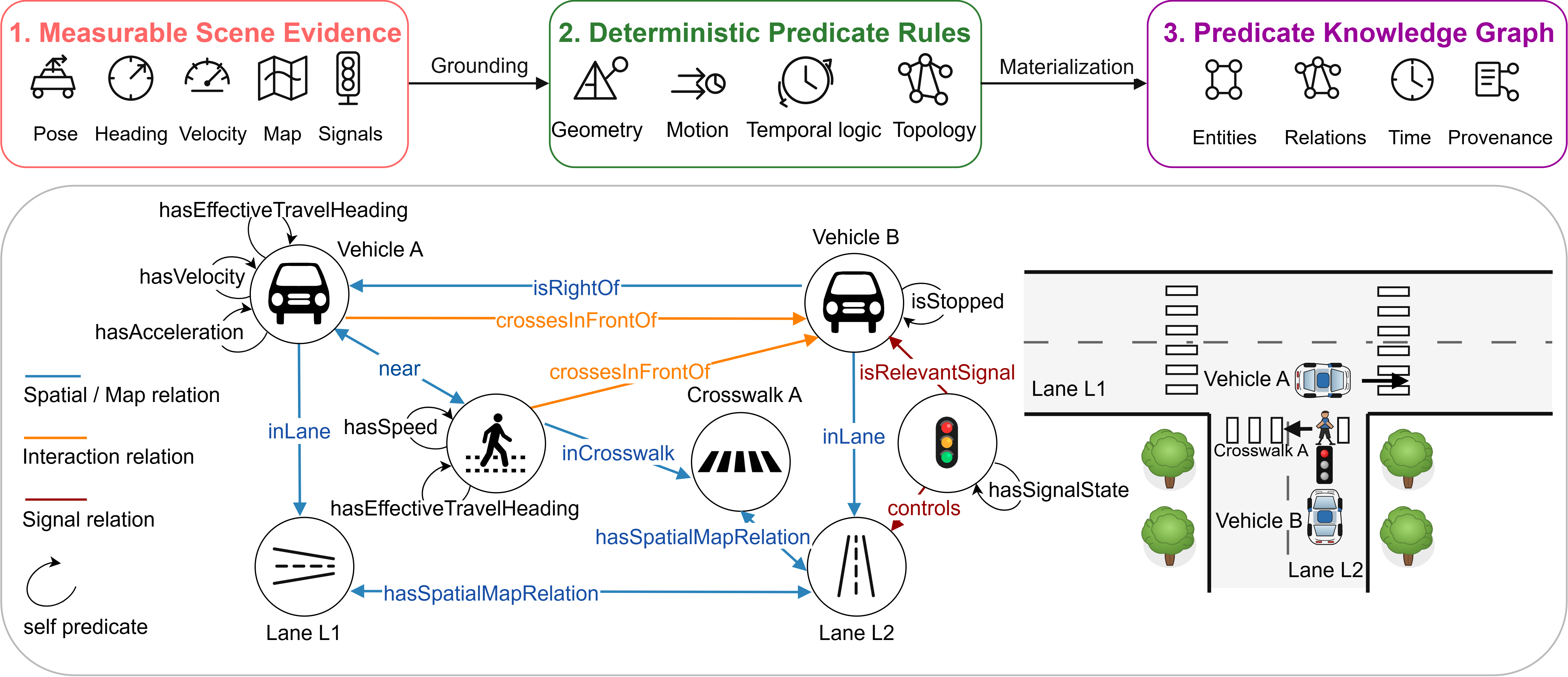}
    \caption{Overview of the proposed framework. Measurable driving-scene evidence is processed through deterministic predicate rules and materialised in a Predicate KG. The example illustrates the correspondence between a physical traffic scene and its semantic predicate representation.}
    \label{fig:framework_overview}
\end{figure*}

To address these limitations, this work introduces a multi-dataset predicate framework for driving-scene reasoning. Semantic relations are derived from measurable geometric, kinematic, temporal, map, traffic-control, and risk-related evidence through explicitly defined and reproducible conditions. Rather than storing dataset annotations in another representation, the Predicate Knowledge Graph (Predicate KG) transforms low-level scene evidence into explicit driving semantics, such as whether one participant follows, yields to, overtakes, or crosses in front of another. Its main purpose is therefore to provide a dataset-independent semantic interface between heterogeneous driving data and downstream reasoning: each assertion can be traced to the underlying scene evidence and deterministic rule, queried independently of the original dataset structure, and directly supplied to reasoning systems such as vision--language models. Figure~\ref{fig:framework_overview} illustrates this process from measurable scene evidence through predicate derivation to the resulting Predicate KG.

The study is guided by the following research questions:

\begin{enumerate}[label=\textbf{RQ\arabic*},ref=RQ\arabic*,leftmargin=*]

\item \label{rq:formalization}
How can driving-scene relations be formalised as deterministic predicates derived from measurable scene information?

\item \label{rq:cross_dataset}
To what extent can a common predicate representation preserve consistent driving semantics across heterogeneous autonomous-driving datasets?

\item \label{rq:vlm_grounding}
How can deterministic predicates be used to ground and evaluate vision--language reasoning against observable scene evidence?

\end{enumerate}

The main contributions of this work are:

\begin{itemize}
    \item A predicate framework that represents driving-scene semantics through explicit and reproducible operational rules applied to measurable scene evidence.
    
    \item A common semantic representation that enables predicates to be derived consistently across nuPlan and nuScenes despite differences in their dataset structure and annotations, while preserving a shared semantic interpretation, and is designed to facilitate extension to additional datasets and downstream applications.    
    
    \item A grounding layer that connects vision--language reasoning with deterministic scene evidence, enabling semantic statements to be traced and evaluated against the underlying traffic situation.
\end{itemize}

\section{Related Work}
\label{sec:state_of_the_art}

Recent advances in vision--language modelling have extended autonomous-driving perception from object-centric recognition toward scene-level interpretation, question answering, explanation, and decision support. DriveGPT4~\cite{xu2024drivegpt4} integrates visual observations with large language models to support interpretable end-to-end driving, while DriveLM~\cite{sima2024drivelm} formulates driving reasoning through graph-structured visual question answering. Complementary benchmarks such as nuScenes-QA~\cite{qian2024nuscenesqa} and LingoQA~\cite{marcu2024lingoqa} evaluate the ability of vision--language models to reason about traffic scenes through natural-language questions and answers. Talk2BEV~\cite{choudhary2024talk2bev} further connects language reasoning with a bird's-eye-view representation of the environment. These approaches demonstrate the potential of language models for semantic driving-scene understanding, but their reasoning remains primarily expressed through learned or generated language, making individual relational statements difficult to verify directly against the underlying scene geometry and dynamics.

This limitation motivates representations that make scene semantics explicit rather than leaving them implicit in model outputs. Ontology- and knowledge-based approaches have long represented traffic participants, infrastructure, and their relations in machine-interpretable form. Buechel et al.~\cite{buechel2017ontology} modelled traffic scenes and regulations using an ontology to support situation awareness and decision-making, while Halilaj et al.~\cite{halilaj2021knowledge} employed knowledge graphs for situation comprehension in driving scenarios. Traffic scene graphs provide a related formulation in which participants are connected through explicit semantic relations; for example, Zipfl et al.~\cite{zipfl2022relation} used spatial semantic relations between traffic participants for motion prediction. The nuScenes Knowledge Graph~\cite{mlodzian2023nusceneskg} demonstrated how rich scene information from a large-scale driving dataset can be transformed into a structured semantic representation.

Explicit structure alone, however, does not guarantee that a relation has a reproducible meaning. For explainable reasoning, predicates such as \textit{inFrontOf}, \textit{leftOf}, \textit{follows}, or \textit{crossesInFrontOf} should be tied to observable geometric, kinematic, temporal, and map conditions rather than inferred only from linguistic context or learned correlations. Rule- and ontology-based approaches provide important foundations for such symbolic reasoning~\cite{buechel2017ontology,halilaj2021knowledge}, while semantic traffic-scene graphs show the value of representing relations explicitly~\cite{zipfl2022relation}. Nevertheless, existing approaches generally employ relations for a particular reasoning, prediction, or representation task rather than defining a broad, deterministic predicate layer intended to serve as verifiable evidence for vision--language reasoning.

A further challenge arises when such semantics are transferred across autonomous-driving datasets. Datasets differ substantially in their object taxonomies, map representations, annotation formats, coordinate systems, and available temporal information. Halilaj et al.~\cite{halilaj2023integration} addressed this heterogeneity through knowledge-graph-based integration of autonomous-driving datasets, demonstrating the value of a shared semantic representation across data sources. However, dataset integration and deterministic semantic grounding address different objectives: the former harmonises heterogeneous information, whereas the latter requires each semantic relation to be computed according to an explicit and reproducible definition. Existing VLM-based driving approaches, structured scene representations, and dataset-integration methods therefore leave an important gap between raw driving data and verifiable language reasoning. Our work addresses this gap through fixed operational predicate definitions that remain unchanged across dataset-specific groundings, are explicitly validated across heterogeneous datasets, and subsequently serve as traceable evidence for vision--language reasoning.

\section{Deterministic Predicate Framework}
\label{sec:framework}

The framework is implemented for two autonomous-driving datasets, nuPlan~\cite{caesar2021nuplan} and nuScenes~\cite{caesar2020nuscenes}. Both datasets provide temporally ordered driving-scene observations with tracked traffic participants and map information, enabling relations to be analysed both at individual time steps and over their temporal evolution. They differ, however, in their data organisation, object taxonomies, map structures, and available annotations. We use dataset-specific grounding only to recover the measurable quantities required by the predicates. Predicate definitions are kept separate from these dataset interfaces, such that the same relation is evaluated according to the same rule in both datasets.

\subsection{Dataset Grounding}
\label{subsec:dataset_grounding}

For each time step $t$, the dataset interface constructs a scene representation containing the observed agents and the available map and traffic-control information. The state of an agent $a_i$ is represented by
\begin{equation}
    \mathbf{s}_i(t) =
    \left[
    \mathbf{p}_i(t),
    \psi_i(t),
    \mathbf{v}_i(t),
    \mathcal{B}_i(t)
    \right],
    \label{eq:agent_state}
\end{equation}
where $\mathbf{p}_i(t)$ is its position, $\psi_i(t)$ its heading, $\mathbf{v}_i(t)$ its velocity, and $\mathcal{B}_i(t)$ its bounding-box dimensions. Map elements and traffic-control information are retained when required by the corresponding predicate.

The role of the dataset interface is limited to obtaining and normalising this evidence. In nuPlan, for example, agent states, lane topology, lane connectors, and traffic-light information are obtained through the corresponding database and map structures. nuScenes exposes comparable scene information through its own sample, annotation, instance, and map records. After this step, predicate computation no longer operates on dataset-specific objects.

\subsection{Predicate Definition and Derivation}
\label{subsec:predicate_derivation}

The predicate vocabulary was designed from semantic relations commonly used in driving ontologies, knowledge graphs, and traffic-scene representations~\cite{buechel2017ontology,halilaj2021knowledge,zipfl2022relation,mlodzian2023nusceneskg}, and was extended with relations required to describe driving interactions and safety-relevant situations. Predicates were retained when they have a meaningful driving interpretation and can be deterministically grounded in measurable or reconstructable scene evidence. The resulting vocabulary is therefore independent of the annotation schema of a particular dataset; nuPlan and nuScenes provide only the evidence required to evaluate the common predicate rules.

Each predicate is associated with an explicit rule over measurable scene information. For a predicate $P$, we write
\begin{equation}
P(\mathbf{x},t)=
\begin{cases}
1, & f_P(\mathbf{x},\mathcal{E}_{\mathcal{W}_P(t)};\boldsymbol{\theta}_P)=\mathrm{true},\\
0, & \mathrm{otherwise},
\end{cases},
\label{eq:predicate_general}
\end{equation}

where $\mathbf{x}$ denotes the participating entity or entities, $\mathcal{E}_{\mathcal{W}_P(t)}$ denotes the scene evidence available within the predicate-specific temporal window $\mathcal{W}_P(t)$, which may consist of the current frame only or multiple observations over time. The function $f_P$ is the predicate-specific Boolean decision function that evaluates whether the operational conditions defining predicate $P$ are satisfied, and $\boldsymbol{\theta}_P$ contains the corresponding rule parameters and threshold functions. Safety-related thresholds for following distance, time gap, collision risk, and lane-change gap are derived from UN Regulation No.~157 (UN R157)~\cite{unece_r157}. The resulting rules are fixed before evaluation and are applied identically to both dataset groundings.

For longitudinal following of M1/N1 vehicles (passenger cars and light goods vehicles up to $3.5\,\mathrm{t}$), the minimum safe following distance $d_{\min}(v)$ is defined using the speed-dependent criterion specified in UN R157. For vehicle speed $v\leq60\,\mathrm{km/h}$,
\begin{equation}
    d_{\min}(v)=v\,t_{\mathrm{front}}(v),
\end{equation}
where $v$ denotes the longitudinal vehicle speed in $\mathrm{m/s}$ and $t_{\mathrm{front}}(v)$ the minimum time gap prescribed by UN R157. The required gap increases piecewise linearly from $1.0\,\mathrm{s}$ at $2.0\,\mathrm{m/s}$ to $1.6\,\mathrm{s}$ at $16.67\,\mathrm{m/s}$ according to the intermediate breakpoints specified in the regulation. For speeds below $2\,\mathrm{m/s}$, a minimum following distance of $2\,\mathrm{m}$ applies. Accordingly, a longitudinal time gap is classified as safe when it is not smaller than $t_{\mathrm{front}}(v)$. Collision-risk conditions follow the UN R157 definition of imminent collision risk: a conflict is classified as imminent when collision avoidance would require a braking demand of at least $5\,\mathrm{m/s^2}$. For regular lane changes, an approaching vehicle in the target lane shall not be required to decelerate at a rate exceeding $3.0\,\mathrm{m/s^2}$ under the specified assessment conditions, while the distance between the two vehicles shall not become smaller than the distance travelled by the Automated Lane Keeping System (ALKS) vehicle in $1.0\,\mathrm{s}$~\cite{unece_r157}. Risk predicates combine relative motion, longitudinal gap, time gap, and collision-avoidance braking demand using these safety criteria.

We distinguish predicates that expose information already available in the dataset from predicates that require an additional derivation. A predicate such as \texttt{hasVelocity}, for instance, retains an observed or reconstructed physical quantity, whereas relations such as \texttt{inFrontOf}, \texttt{follows}, \texttt{overtakes}, and \texttt{crossesInFrontOf} require geometric, kinematic, temporal, or map conditions to be evaluated. This distinction is also reflected in the validation described in Sec.~\ref{sec:evaluation}, where the main evaluation concerns the derived predicates rather than quantities obtained directly from the source data.

The predicate inventory comprises spatial, motion, temporal, map, interaction, traffic-control, and risk families. Table~\ref{tab:predicate_groups} summarises representative higher-level predicates; complete implementation-grounded definitions and thresholds are provided in the \hyperref[app:predicate_definitions]{Appendix}.

\begin{table*}[t]
\centering
\caption{Operational summary of the evaluated higher-level predicates.}
\label{tab:predicate_groups}
\begin{tabular}{lp{0.68\textwidth}}
\toprule
\textbf{Predicate} & \textbf{Operational summary} \\
\midrule
\texttt{follows}, \texttt{queuesBehind} &
Persistent longitudinal relation between participants on a compatible travel path;\newline
\texttt{queuesBehind} additionally captures queued traffic. \\

\texttt{changesLane} &
Temporal transition of a participant between lane memberships during a lane-change manoeuvre. \\

\texttt{mergesInFrontOf}, \texttt{mergesBehind} &
Merge into a common travel path, distinguished by the resulting longitudinal ordering of the participants. \\

\texttt{overtakes} &
Temporal manoeuvre in which the relative ordering changes from behind to ahead. \\

\texttt{yieldsTo} &
Interaction in which a participant gives precedence to another participant during a conflicting movement. \\
\bottomrule
\end{tabular}
\end{table*}

Spatial predicates are computed relative to the orientation of the reference participant. For a subject agent $S$ and an object agent $O$, their relative position is
\begin{equation}
    \Delta\mathbf{p}_{SO}(t)=\mathbf{p}_{O}(t)-\mathbf{p}_{S}(t).
\end{equation}
Its longitudinal and lateral components in the local coordinate system of $S$ are used to evaluate relations such as \texttt{inFrontOf}, \texttt{leftOf}, and \texttt{rightOf}. This avoids defining directional relations with respect to a fixed global map orientation.

Temporal and interaction predicates combine information from several state variables or time steps. As one representative example, \texttt{crossesInFrontOf} evaluates whether the forward motion directions of two participants lead to a common conflict point. For a subject $S$ and object $O$, the forward rays are defined as
\begin{align}
    R_S(\lambda) &= \mathbf{p}_S + \lambda\mathbf{d}_S, \qquad 0 \leq \lambda \leq L, \\
    R_O(\mu) &= \mathbf{p}_O + \mu\mathbf{d}_O, \qquad 0 \leq \mu \leq L,
\end{align}
where $\mathbf{d}_S$ and $\mathbf{d}_O$ are their directions of travel and $L$ defines the forward range. In the current implementation, $L=10\,\mathrm{m}$. An intersection is considered only when it lies in the forward direction of both participants and satisfies the remaining motion and contextual conditions of the predicate. The finite forward range prevents intersections of distant trajectory lines from being interpreted as local crossing events.

Map predicates combine participant geometry with the HD map rather than relying solely on the agent centre. This is relevant near lane boundaries and intersections, where a participant footprint may already occupy a map element while its centre remains outside.

Traffic-control semantics are represented through separate predicates. The relation
\begin{equation}
    \texttt{controls}(\sigma,m)
\end{equation}
links a traffic signal $\sigma$ to the lane or lane-connector movement $m$ whose traffic flow is governed by that signal.

\begin{equation}
    \texttt{hasSignalState}(\sigma,q)
\end{equation}
records its current signal state $q$. The agent-specific relation
\begin{equation}
\texttt{isRelevantSignal}(a,\sigma)
\end{equation}
is asserted when the traffic signal $\sigma$ controls a lane or lane connector on the approaching path of agent $a$. It is defined as a pre-entry relation and is no longer emitted once the agent begins to occupy the controlled lane connector. Keeping these predicates separate preserves the distinction between the traffic signal, its current state, the movement it controls, and the approaching participant affected by that control.

\subsection{Predicate Knowledge Graph and Implementation}
\label{subsec:predicate_kg}

A derived relation is stored together with the participating entities, its temporal reference, and provenance information. A binary assertion is represented conceptually as
\begin{equation}
e=\left(S,P,O,t,\pi\right),
\label{eq:predicate_assertion}
\end{equation}
where $S$ and $O$ denote the subject and object, $P$ the predicate, and $t$ the corresponding time. The provenance information $\pi$ records the origin of the assertion, including the source scene evidence and the predicate rule from which it was derived. Unary predicates follow the same representation without a relational object. The resulting assertions are materialised in the Predicate KG and can subsequently be queried without requiring knowledge of the original nuPlan or nuScenes data organisation.

To provide an indication of the scale of the resulting representation, a representative nuPlan scenario comprising 41 frames contains 741,311 materialised predicate assertions, corresponding to approximately 18,081 assertions per frame. In one inspected frame, the Predicate KG contains 89 dynamic entity nodes and 4,752 semantic edges, representing 21,357 predicate assertions. These assertions cover multiple relation directions and entity combinations, including ego--agent, agent--ego, agent--agent, ego--structure, structure--ego, agent--structure, and structure--agent relations. Multiple predicates may describe the same entity pair, explaining the difference between semantic edges and predicate assertions. Depending on the downstream task, the graph can be filtered to retain only the predicate families, entity types, or relation directions relevant to the analysis.

The implementation separates dataset grounding, predicate computation, and graph materialisation. \hyperref[app:predicate_definitions]{ The Appendix} provides the implementation-grounded predicate definitions, including their required evidence, mathematical formulation, applicability conditions, and thresholds. The corresponding implementation and predicate specification are available in the repository provided in Sec.~\ref{sec:code_availability}.

\section{Experimental Evaluation}
\label{sec:evaluation}

The evaluation focuses on the derived predicates defined in Sec.~\ref{subsec:predicate_derivation}. Validation combines programmatic checks with semantic inspection of the underlying traffic scenes. Temporal context is also considered when a predicate's interpretation depends on the evolution of an interaction.

\subsection{Automatic Validation}
\label{subsec:automatic_validation}

Each generated Predicate KG is checked for valid entities and arguments, finite numerical values, temporal continuity, predicate dependencies, mutually exclusive relations, and rule-specific invariants. Numerical quantities are independently recomputed where applicable, and regression tests cover directed spatial relations, motion and temporal histories, map associations, interaction, traffic-control, and risk predicates. For nuPlan, map-dependent predicates are additionally verified against the native HD map by reconstructing agent footprints and recomputing lane and lane-connector membership, intersection association, longitudinal progress, and map heading without reusing the generated assertions.

Predicate rules and thresholds are fixed before semantic evaluation and are not tuned on the validation scenarios. For \texttt{crossesInFrontOf}, a preliminary sensitivity analysis on separate development scenarios with $L\in\{5,10,15\}\,\mathrm{m}$ selected $L=10\,\mathrm{m}$, which was then fixed for both datasets.

\subsection{Semantic and Temporal Validation}
\label{subsec:semantic_validation}

Semantic validation is conducted on 200 scenarios from each dataset. The quantitative evaluation focuses on higher-level interaction and traffic-control predicates, while spatial, motion, map, and risk predicates are additionally examined through rule-specific consistency checks and visual inspection. nuPlan scenarios are selected using the official \texttt{scenario\_tag} annotations, whereas comparable nuScenes categories are constructed from scene metadata, map context, and object annotations. The selected scenarios cover \emph{intersection traversal}, \emph{lane following}, \emph{lane changes}, \emph{merging}, \emph{overtaking}, \emph{crossing interactions}, \emph{traffic-light approaches}, \emph{pedestrian and cyclist interactions}, \emph{stationary and moving traffic}, and \emph{dense multi-agent scenes}. Scenario selection is fixed before inspection of predicate correctness. The scenario is used as the sampling unit because several predicate instances may occur simultaneously among different participants and map elements.

The validation visualisation (a BEV-based inspection interface) contains the complete bird's-eye-view scene, agent footprints, HD-map context, and the generated relations. Spatial, map, and risk predicates are inspected at frame level or over the required local temporal context, while higher-level interactions are reviewed over their supporting temporal sequence. In particular, \texttt{changesLane}, \texttt{mergesInFrontOf}, \texttt{mergesBehind}, \texttt{overtakes}, and \texttt{yieldsTo} are assessed over the corresponding manoeuvre evolution rather than from isolated frames. Persistent relations such as \texttt{follows} and \texttt{queuesBehind} are inspected over consecutive observations.

For \texttt{crossesInFrontOf}, subsequently observed trajectories provide a retrospective check of whether the detected local conflict is consistent with the realised motion of the participants. Future observations and observed-future predicates are restricted to validation and are not provided as inputs to the vision--language experiments in Sec.~\ref{sec:vlm_reasoning}.

For each evaluated predicate, candidates are generated independently of the predicate output from entity relations satisfying its applicability conditions. Both asserted and non-asserted candidates are reviewed, enabling false positives and false negatives to be identified without treating arbitrary missing graph edges as negatives. Each candidate is manually labelled as \emph{present}, \emph{absent}, or \emph{ambiguous} based on the BEV scene, HD-map context, and, where required, its temporal sequence. Ambiguous cases are excluded from the computation of precision, recall, and $F_1$ and are recorded separately. Across the reviewed instances, ambiguous cases account for 2.92\% in nuPlan and 2.45\% in nuScenes.

Quantitative semantic validation is performed on reviewed candidate higher-level instances rather than on all frame-level predicate assertions. Across the 200 scenarios, this results in 1,607 reviewed higher-level instances for nuPlan and 1,266 for nuScenes. Of the nuPlan instances, 1,427 belong to the eight predicates supported by both datasets, while 180 correspond to \texttt{isRelevantSignal}. Table~\ref{tab:predicate_validation} reports the number of reviewed candidates and the resulting $F_1$ scores. Predicate-level precision, recall, and ambiguity statistics are retained with the complete validation results.

\begin{table}[t]
\centering
\caption{Semantic validation of higher-level derived predicates. $N$ denotes the number of reviewed candidates before exclusion of ambiguous cases.}
\label{tab:predicate_validation}
\begin{tabularx}{\linewidth}{Xccccc}
\toprule
& \multicolumn{2}{c}{\textbf{nuPlan}} & \multicolumn{2}{c}{\textbf{nuScenes}} & \\
\cmidrule{2-3} \cmidrule{4-5}
\textbf{Predicate} & $\boldsymbol{N}$ & $\boldsymbol{F_1}$ & $\boldsymbol{N}$ & $\boldsymbol{F_1}$ & $\boldsymbol{\Delta F_1}$ \\
\midrule
\texttt{follows}           & 470 & 0.98 & 430 & 0.97 & 0.01 \\
\texttt{queuesBehind}      & 260 & 0.99 & 220 & 0.99 & 0.00 \\
\texttt{changesLane}       & 165 & 0.92 & 150 & 0.95 & 0.03 \\
\texttt{mergesInFrontOf}   & 105 & 0.96 &  95 & 0.93 & 0.03 \\
\texttt{mergesBehind}      &  85 & 0.94 &  75 & 0.93 & 0.01 \\
\texttt{overtakes}         &  62 & 0.93 &  51 & 0.87 & 0.06 \\
\texttt{crossesInFrontOf}  & 165 & 0.90 & 145 & 0.92 & 0.02 \\
\texttt{yieldsTo}          & 115 & 0.88 & 100 & 0.90 & 0.02 \\
\midrule
\texttt{isRelevantSignal}  & 180 & 0.95 & -- & -- & -- \\
\midrule
\textbf{Macro average}     & -- & \textbf{0.94} & -- & \textbf{0.93} & \textbf{0.02} \\
\bottomrule
\end{tabularx}
\end{table}

The results show high agreement between the rule-derived predicates and the manual scene-based labels, with macro $F_1$ scores of 0.94 on nuPlan and 0.93 on nuScenes. \texttt{follows} and \texttt{queuesBehind} obtain the highest agreement, whereas relations involving more extended temporal evolution show greater variation. The largest cross-dataset difference is observed for \texttt{overtakes}, with $F_1=0.93$ on nuPlan and $0.87$ on nuScenes. This difference is consistent with the stronger dependence of overtaking on temporal continuity, lane transitions, and completion of the manoeuvre. \texttt{crossesInFrontOf} and \texttt{yieldsTo} reach $F_1$ scores between 0.88 and 0.92 despite requiring the joint interpretation of geometry, motion, and temporal context.

For the eight predicates supported by both datasets, the mean absolute cross-dataset difference is $\Delta F_1=0.02$, and seven predicates differ by no more than 0.03. Errors correspond to cases where the rule-derived predicate disagrees with the manual scene-based label, for example because a threshold or applicability condition does not fully capture the observed situation. The comparable aggregate performance, together with the predicate-specific differences, shows that the operational definitions remain stable under the two dataset groundings. \texttt{isRelevantSignal} achieves an $F_1$ score of 0.95 on nuPlan but is excluded from the cross-dataset comparison because equivalent traffic-control evidence is not available in the current nuScenes grounding.

\section{Predicate-Grounded Vision--Language Reasoning}
\label{sec:vlm_reasoning}

To examine whether the Predicate KG provides useful grounding for vision--language reasoning, we follow the multi-frame NuPlanQA-Eval setting~\cite{park2025nuplanqa} using a frozen LLaVA-OneVision-7B model. No additional training or fine-tuning is performed, and the questions and decoding configuration are kept fixed across all conditions. Five input conditions are evaluated: the original visual-history baseline; visual input augmented with raw ground-truth metadata; visual input augmented with the serialised Predicate KG; and two corresponding no-visual conditions using either metadata or the Predicate KG as the scene representation. The Predicate KG input additionally includes a fixed glossary defining predicate semantics, while mathematical rules and thresholds are not exposed to the VLM. Raw metadata contains the primitive quantities used for predicate derivation, including agent states, motion, map associations, and traffic-control information, but no derived semantic relations.

The raw metadata and predicate conditions are derived from the same entities, timestamps, and underlying temporal evidence, but expose this information at different semantic abstraction levels. Predicate selection is not random and does not depend on the question. Instead, a fixed scene-based selection procedure is applied identically to all samples, retaining the available predicate relations for the entities represented in the corresponding visual history. Metadata and predicate inputs use only observations up to the final timestamp of the visual history; future observations and retrospective validation predicates are excluded. The resulting temporally available Predicate KG subgraph is serialised into compact assertions such as \texttt{crossesInFrontOf(pedestrian\_4, ego)}, \texttt{isRelevantSignal(ego, signal\_7)}, and \texttt{hasSignalState(signal\_7, RED)}.

Because the metadata and Predicate KG conditions originate from annotated nuPlan states and map information, the experiment evaluates oracle semantic grounding rather than visual predicate prediction. Removing the visual history provides a complementary comparison of how much reasoning can be supported by these scene representations independently of the image sequence.

We report four-option accuracy (25\% chance) on the 1,801-question NuPlanQA-Eval test set across nine subtasks.

\begin{table*}[t]
\centering
\caption{Effect of structured scene grounding on the nine NuPlanQA-Eval subtasks.
Bold indicates the best result among conditions with visual input; underlining
indicates the best result without visual input. All values are accuracy (\%).}
\label{tab:vlm_grounding}

%\scriptsize
%\setlength{\tabcolsep}{4.5pt}

\begin{tabularx}{\linewidth}{Xccccccccc}
%\hline
\toprule
&
\multicolumn{3}{c}{\textbf{Road Env. Perception}} &
\multicolumn{3}{c}{\textbf{Spatial Relations Recog.}} &
\multicolumn{3}{c}{\textbf{Ego-centric Reasoning}} \\
\cmidrule{2-4} \cmidrule{5-7} \cmidrule{8-10}
%\noalign{\vskip 2pt}
\textbf{Method} &
\textit{Traffic Light} &
\textit{Weather/Lighting} &
\textit{Road Type} &
\textit{Sur. Obj.} &
\textit{Traffic Flow} &
\textit{Key Obj.} &
\textit{Ego Ctrl.} &
\textit{Situ. Asse.} &
\textit{Act. Rec.} \\
\midrule
%\hline
%\noalign{\vskip 1pt}

\multicolumn{10}{c}{\textbf{Visual input}} \\
\midrule
%\hline
%\noalign{\vskip 2pt}

LLaVA-OV-7B
& 53.2
& \textbf{89.4}
& 96.4
& 77.9
& 77.6
& 73.2
& 75.4
& 76.2
& 82.9 \\

+ GT Metadata
& 62.8
& 89.0
& 96.5
& \textbf{83.0}
& 81.0
& 77.5
& 63.4
& 81.0
& 84.0 \\

+ Predicate KG
& \textbf{71.5}
& 88.8
& \textbf{96.8}
& 82.7
& \textbf{85.8}
& \textbf{82.3}
& \textbf{83.9}
& \textbf{86.1}
& \textbf{89.0} \\
\midrule
%\hline
%\noalign{\vskip 2pt}
\multicolumn{10}{c}{\textbf{No visual input}} \\
\midrule
%\noalign{\vskip 2pt}

GT Metadata Only
& 47.6
& \underline{26.8}
& 71.9
& 75.1
& 68.7
& 65.8
& 60.8
& 69.3
& 66.7 \\

Predicate KG Only
& \underline{64.2}
& 25.6
& \underline{79.1}
& \underline{78.8}
& \underline{78.4}
& \underline{74.6}
& \underline{73.1}
& \underline{79.3}
& \underline{76.8} \\

\bottomrule
\end{tabularx}
\end{table*}

Table~\ref{tab:vlm_grounding} shows that predicate grounding achieves the highest accuracy in seven of nine visual-input comparisons. Large gains are observed for Traffic Light (53.2\% to 71.5\%), Traffic Flow (77.6\% to 85.8\%), Ego-Vehicle Manoeuvre (75.4\% to 83.9\%), Situation Assessment (76.2\% to 86.1\%), and Action Recommendation (82.9\% to 89.0\%). Unlike raw metadata, the Predicate KG combines geometric, kinematic, temporal, map, and traffic-control evidence through predefined rules with regulation-derived or empirically validated thresholds, exposing higher-level relations that the VLM would otherwise need to infer from individual measurements. This is particularly visible for Ego-Vehicle Manoeuvre, where raw metadata reaches 63.4\% while predicate grounding reaches 83.9\%. Surrounding Objects is the only subtask where raw metadata performs best (83.0\% vs.\ 82.7\%), while Weather/Lighting remains essentially unchanged (89.4\% vs.\ 88.8\%), consistent with the absence of weather predicates. Because both conditions originate from the same underlying scene evidence, these gains indicate that the explicit semantic abstraction provided by the Predicate KG is useful for downstream reasoning, rather than merely providing the model with additional information.

Without visual input, the Predicate KG outperforms raw metadata in eight of nine subtasks, including Traffic Light (64.2\% vs.\ 47.6\%), Traffic Flow (78.4\% vs.\ 68.7\%), Ego-Vehicle Manoeuvre (73.1\% vs.\ 60.8\%), Situation Assessment (79.3\% vs.\ 69.3\%), and Action Recommendation (76.8\% vs.\ 66.7\%). Notably, Predicate KG only input also matches or exceeds the visual baseline in five subtasks, including Traffic Light, Surrounding Objects, Traffic Flow, Key Objects, and Situation Assessment. This shows that, once structured scene semantics are available, several reasoning tasks can be supported without providing the image sequence directly to the VLM. Weather/Lighting is the exception and falls close to the 25\% random-choice level (25.6\%), confirming that this benefit is limited to information represented by predicates.

Figure~\ref{fig:predicate_examples} illustrates representative predicate relations at three selected time steps, $t_1<t_2<t_3$, whose temporal spacing may vary across examples. For space reasons, the figure presents only three representative predicate families: spatial, interaction, and traffic-control. The spatial row focuses on ego--agent relations, while the interaction row additionally includes agent--agent relations. The traffic-control row shows signal relevance, controlled movements, and signal states. Together, these examples illustrate how predicate relations evolve over time.

\section{Discussion and Limitations}
\label{sec:discussion}

The cross-dataset results indicate that fixed predicate definitions preserve comparable semantics across the two dataset groundings, although temporally extended relations such as \texttt{overtakes} remain more sensitive to trajectory continuity and annotation characteristics.

In NuPlanQA, predicate grounding primarily benefits interaction- and reasoning-oriented tasks; benefits are smaller for information directly available from object annotations or absent from the predicate vocabulary.

The framework is constrained by the evidence available in each dataset. A common vocabulary does not imply that every predicate can be instantiated for every source. In the current implementation, for example, \texttt{isRelevantSignal} is evaluated in nuPlan but excluded from the cross-dataset comparison because equivalent traffic-control evidence is not available in the current nuScenes grounding. Predicate reliability also depends on the quality of object tracks, map geometry, temporal coverage, and the assumptions underlying the predicate-specific threshold functions. The risk family is therefore grounded in explicit safety-related quantities and thresholds rather than learned or dataset-specific semantic labels. Although the evaluated predicates show high agreement in the considered scenarios, rare or weakly represented interactions may require broader validation.

Beyond VLM grounding, the Predicate KG can support structured querying and consistency checking against measurable scene evidence. Future work will investigate predicate extraction directly from sensor observations, connecting the oracle-grounding setting to a perception-driven pipeline.

% Width of every image panel -- change only this value
\newcommand{\predfigwidth}{0.28\textwidth}

% Width reserved for the vertical family name
\newcommand{\predlabelwidth}{0.026\textwidth}

\begin{figure*}[t]
    \centering

    \noindent
    \begin{minipage}[c]{\predlabelwidth}
    \end{minipage}
    \hfill
    \begin{minipage}[c]{\predfigwidth}
        \centering \textbf{$t_1$}
    \end{minipage}
    \hfill
    \begin{minipage}[c]{\predfigwidth}
        \centering \textbf{$t_2$}
    \end{minipage}
    \hfill
    \begin{minipage}[c]{\predfigwidth}
        \centering \textbf{$t_3$}
    \end{minipage}

    \vspace{1mm}

    % =========================================================
    % Spatial
    % =========================================================
    \begin{minipage}[c]{\predlabelwidth}
        \centering
        \rotatebox[origin=c]{90}{\textbf{\small Spatial}}
    \end{minipage}
    \hfill
    \begin{subfigure}[c]{\predfigwidth}
        \centering
        \begin{tikzpicture}
            \node[inner sep=0] (img) {\includegraphics[width=\linewidth]{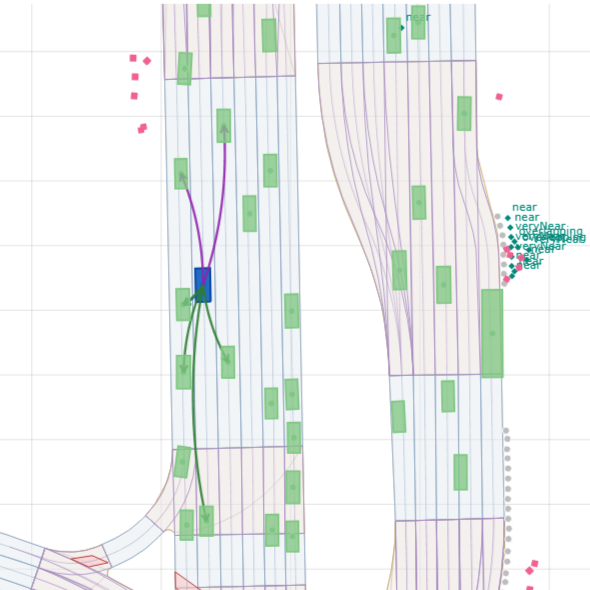}};
            \draw[black, line width=1.4pt, -{Stealth[length=2.5mm,width=2mm]}]
                ($(img.north east)+(-6mm,-4mm)$) -- ($(img.north east)+(-6mm,-16mm)$);
        \end{tikzpicture}
    \end{subfigure}
    \hfill
    \begin{subfigure}[c]{\predfigwidth}
        \centering
        \includegraphics[width=\linewidth]{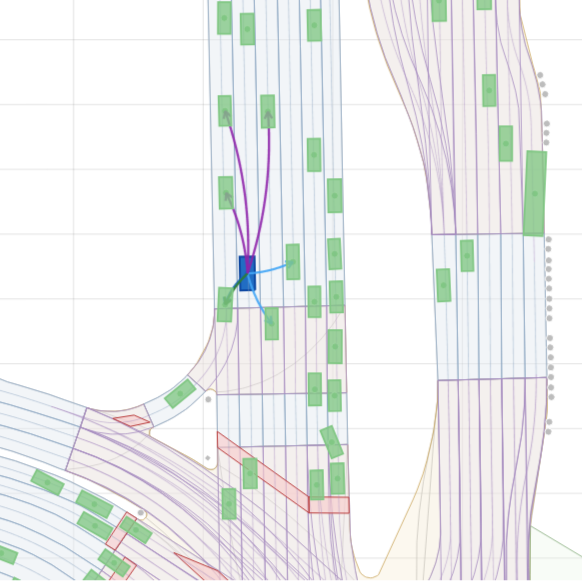}
    \end{subfigure}
    \hfill
    \begin{subfigure}[c]{\predfigwidth}
        \centering
        \includegraphics[width=\linewidth]{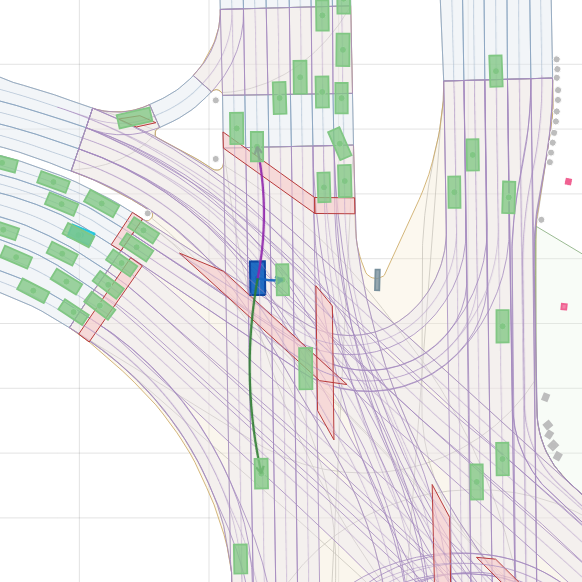}
    \end{subfigure}

    \vspace{0.5mm}

    {\scriptsize
    \legendline{cInFront}{inFrontOf}\quad
    \legendline{cBehind}{behind}\quad
    \legendline{cLeft}{leftOf}\quad
    \legendline{cFrontLeft}{frontLeftOf}\quad
    }

    \vspace{1.5mm}

    % =========================================================
    % Interaction
    % =========================================================
    \begin{minipage}[c]{\predlabelwidth}
        \centering
        \rotatebox[origin=c]{90}{\textbf{\small Interaction}}
    \end{minipage}
    \hfill
    \begin{subfigure}[c]{\predfigwidth}
        \centering
        \begin{tikzpicture}
            \node[inner sep=0] (img) {\includegraphics[width=\linewidth]{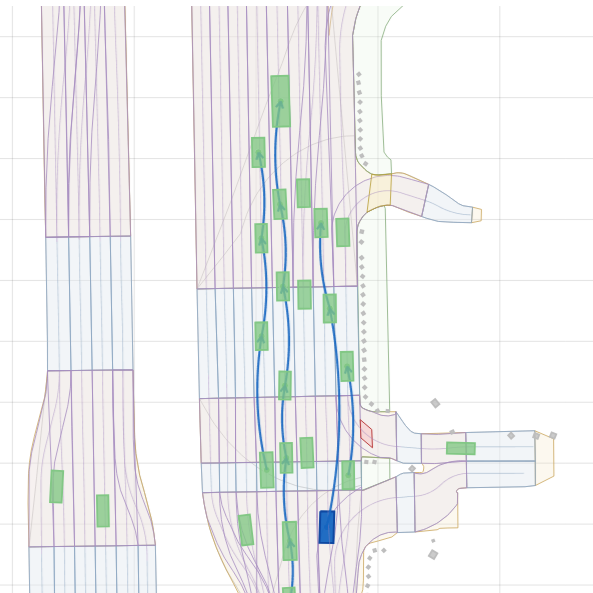}};
            \draw[black, line width=1.4pt, -{Stealth[length=2.5mm,width=2mm]}]
            ($(img.north east)+(-6mm,-16mm)$) -- ($(img.north east)+(-6mm,-4mm)$);
        \end{tikzpicture}
    \end{subfigure}
    \hfill
    \begin{subfigure}[c]{\predfigwidth}
        \centering
        \includegraphics[width=\linewidth]{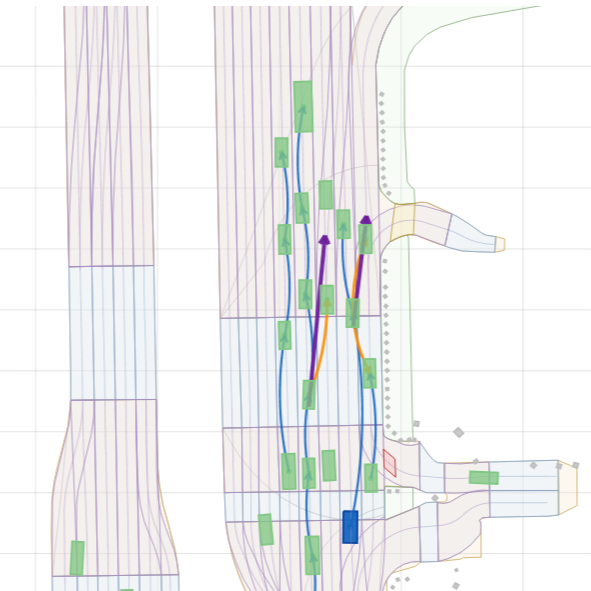}
    \end{subfigure}
    \hfill
    \begin{subfigure}[c]{\predfigwidth}
        \centering
        \includegraphics[width=\linewidth]{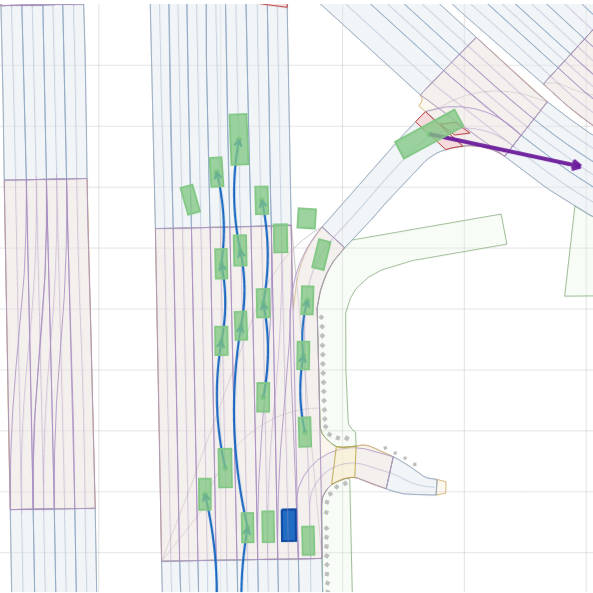}
    \end{subfigure}

    \vspace{0.5mm}

    {\scriptsize
    \legendline{cFollows}{follows}\quad
    \legendline{cMerge}{merge}\quad
    \legendline{cLaneChange}{changesLane}
    }

    \vspace{1.5mm}

    % =========================================================
    % Traffic Signs
    % =========================================================
    \begin{minipage}[c]{\predlabelwidth}
        \centering
        \rotatebox[origin=c]{90}{\textbf{\small Traffic-control}}
    \end{minipage}
    \hfill
        \begin{subfigure}[c]{\predfigwidth}
        \centering
        \begin{tikzpicture}
            \node[inner sep=0] (img) {\includegraphics[width=\linewidth]{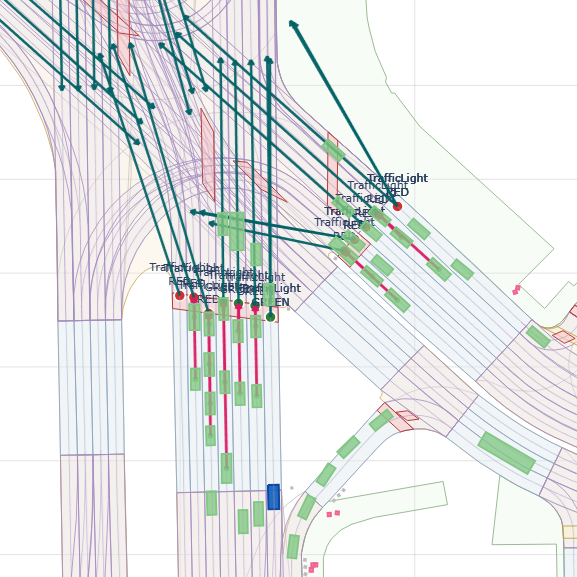}};
            \draw[black, line width=1.4pt, -{Stealth[length=2.5mm,width=2mm]}]
            ($(img.north east)+(-6mm,-16mm)$) -- ($(img.north east)+(-6mm,-4mm)$);
        \end{tikzpicture}
    \end{subfigure}
    \hfill
    \begin{subfigure}[c]{\predfigwidth}
        \centering
        \includegraphics[width=\linewidth]{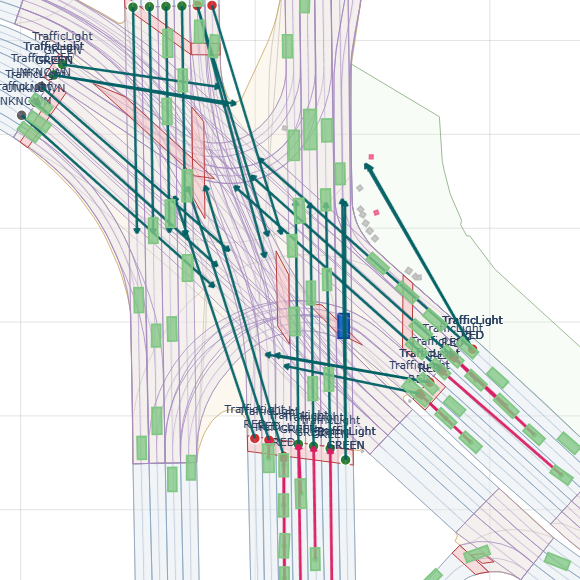}
    \end{subfigure}
    \hfill
    \begin{subfigure}[c]{\predfigwidth}
        \centering
        \includegraphics[width=\linewidth]{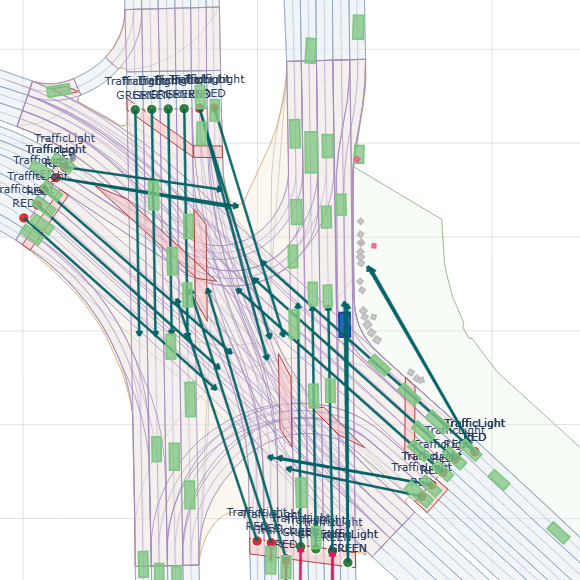}
    \end{subfigure}

    \vspace{0.5mm}

    {\scriptsize
    \legendline{cTLRelevant}{isRelevantSignal}\quad
    \legendline{cTLControls}{controls}\quad
    \halfRGcircle\,hasSignalState
    }

    \caption{Representative predicate relations at three selected time steps for three predicate families shown as illustrative examples: spatial, interaction, and traffic-control. Colours denote predicate semantics, and the black arrow indicates the ego vehicle's travel direction.}

    \label{fig:predicate_examples}
\end{figure*}
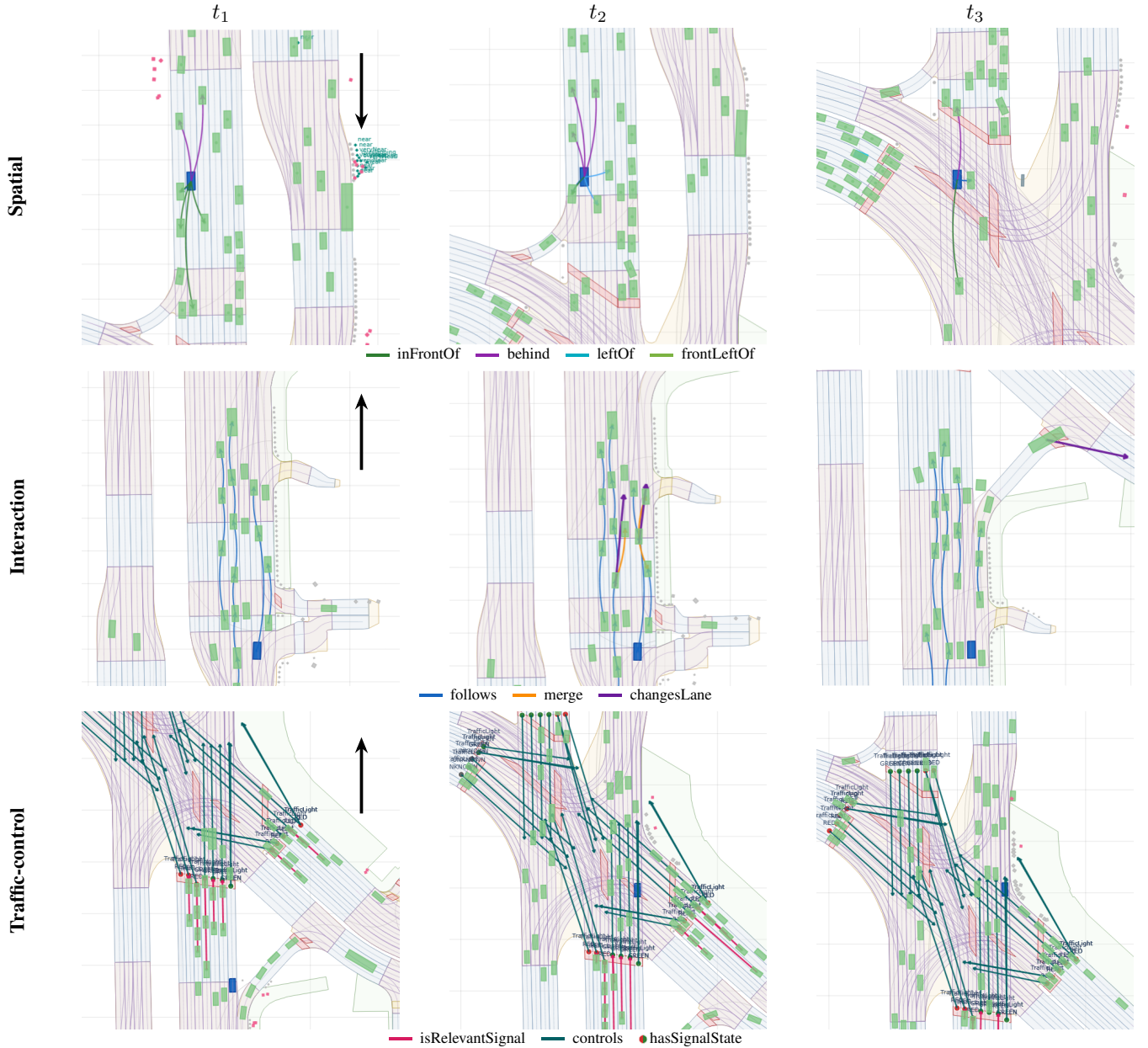

\section{Code Availability}
\label{sec:code_availability}

The implementation and repository version of the predicate specification are available at:
\url{https://github.com/AImotion-Bavaria/nuPlan_Predicates_KG.git}

\section{Conclusion}
\label{sec:conclusion}
This paper introduced a multi-dataset predicate framework for grounding driving-scene semantics in measurable geometric, kinematic, temporal, map, and traffic-control evidence. A common predicate representation separates dataset-specific grounding from the semantic rules used to derive relations. Validation on nuPlan and nuScenes yielded macro $F_1$ scores of 0.94 and 0.93, respectively, demonstrating reproducible derivation of higher-level driving relations from observable scene information (\ref{rq:formalization}). Across the two datasets, the shared predicates differed by only 0.02 in $F_1$ on average, with seven of eight predicates differing by no more than 0.03 (\ref{rq:cross_dataset}). We further used the Predicate KG to ground a frozen vision--language model on NuPlanQA, achieving the highest accuracy among the evaluated visual-input conditions in seven of nine subtasks, while Predicate KG only input outperformed metadata-only input in eight of nine subtasks (\ref{rq:vlm_grounding}). These findings show that, under oracle grounding, predicates can provide a consistent and traceable semantic layer between driving data and vision--language reasoning. Future work will extend dataset coverage and investigate predicate extraction directly from sensor observations.

\appendix
\section{Predicate Definitions}
\label{app:predicate_definitions}

This appendix provides the implementation-grounded operational definitions of the semantic predicates used in this work. The predicates are grouped into the seven Predicate KG families: spatial, motion, temporal, map, interaction, traffic-control, and risk.

The appendix covers 83 implemented semantic predicates: 12 spatial, 17 motion, 17 temporal, 25 map, 8 interaction, 3 traffic-control, and 1 risk predicate.

Throughout,
\begin{equation}
    P\defiff C
\end{equation}
denotes a definitional equivalence: predicate $P$ is defined to hold exactly when condition $C$ is satisfied. Ordinary equalities are used for stored numerical quantities. Numerical thresholds are stated with the corresponding predicate definitions or, for safety-related criteria, in Appendix~\ref{appsec:safety_thresholds}.

\subsection{Common Notation}
\label{appsec:notation}

Let $E$ denote a spatial entity, $S$ the subject of a directed pair, and $O$ the object. Their planar centers are $\mathbf{p}_S=[x_S,y_S]^\top$ and $\mathbf{p}_O=[x_O,y_O]^\top$, with subject body heading $\theta_S$. The relative position is
\begin{equation}
    \Delta\mathbf{p}_{SO}
    =\mathbf{p}_O-\mathbf{p}_S=[dx,dy]^\top .
\end{equation}
The subject-forward and subject-left unit vectors are
\begin{equation}
\mathbf{e}_S=
\begin{bmatrix}\cos\theta_S\\\sin\theta_S\end{bmatrix},
\qquad
\mathbf{n}_S=
\begin{bmatrix}-\sin\theta_S\\\cos\theta_S\end{bmatrix}.
\end{equation}
The body-frame longitudinal and lateral displacements are
\begin{align}
    l_{SO}&=\mathbf{e}_S^\top\Delta\mathbf{p}_{SO},\\
    r_{SO}&=\mathbf{n}_S^\top\Delta\mathbf{p}_{SO},
\end{align}
so that $l_{SO}>0$ means ahead and $r_{SO}>0$ means left. The center distance is
\begin{equation}
    d_c(S,O)=\|\Delta\mathbf{p}_{SO}\|_2.
\end{equation}

Let $F_E$ be the oriented rectangular footprint of entity $E$. The free-space clearance and intersection area of a pair are
\begin{align}
d_{fs}(S,O)&=\operatorname{dist}(F_S,F_O),\\
A_\cap(S,O)&=\operatorname{area}(F_S\cap F_O).
\end{align}
Let $\mathbf{v}_S$ and $\mathbf{v}_O$ be their global planar velocity vectors and
\begin{equation}
    \Delta\mathbf{v}_{SO}=\mathbf{v}_O-\mathbf{v}_S
    =[\Delta v_x,\Delta v_y]^\top.
\end{equation}
The operator $\operatorname{wrap}(\alpha)$ maps an angle to $[-\pi,\pi)$. For temporal predicates, $k$ indexes an observation and $t_k$ denotes its timestamp. Dataset-specific timestamps are normalised to seconds before predicate evaluation, such that
\begin{equation}
    \Delta t_k=t_k-t_{k-1}
\end{equation}
is expressed in seconds. With the default sample interval $T_s=0.5\,\mathrm{s}$ and maximum temporal-gap factor $1.5$, continuity is preserved only when $\Delta t_k\leq0.75\,\mathrm{s}$.

%%%%%%%%%%%%%%%%%%%%%%%%%%%%%%%%%%%%%%%%%%%%%%%%%%%%%%%%%%%%%%%%%%%%%%%%%%%%%%%%

\subsection{Spatial Predicates}
\label{appsec:spatial}

The eight directional predicates are classified in the subject body-heading frame. The defaults are a longitudinal deadband $\tau_l=1.0\,\mathrm{m}$, lateral deadband $\tau_r=0.75\,\mathrm{m}$, pure-left/right band $\tau_c=2.0\,\mathrm{m}$, and front/rear corridor
\begin{equation}
    w(l)=2.5+0.30|l|\quad[\mathrm{m}].
\end{equation}
Exactly one primary directional sector is selected when the pair geometry is valid.

\paragraph{\pred{inFrontOf}}
\begin{equation}
\begin{aligned}
\pred{inFrontOf}(S,O)\defiff{}& l_{SO}\geq1\land\Bigl[\\
&(|l_{SO}|\leq2\land |r_{SO}|<0.75)\\
&\lor\bigl(|l_{SO}|>2\\
&\qquad\land |r_{SO}|\leq w(l_{SO})\bigr)\Bigr].
\end{aligned}
\end{equation}

\paragraph{\pred{behind}}
\begin{equation}
\begin{aligned}
\pred{behind}(S,O)\defiff{}& l_{SO}\leq-1\land\Bigl[\\
&(|l_{SO}|\leq2\land |r_{SO}|<0.75)\\
&\lor\bigl(|l_{SO}|>2\\
&\qquad\land |r_{SO}|\leq w(l_{SO})\bigr)\Bigr].
\end{aligned}
\end{equation}

\paragraph{\pred{leftOf} and \pred{rightOf}}
\begin{align}
\pred{leftOf}(S,O)
&\defiff |l_{SO}|\leq2\land r_{SO}\geq0.75,\\
\pred{rightOf}(S,O)
&\defiff |l_{SO}|\leq2\land r_{SO}\leq-0.75.
\end{align}

\paragraph{\pred{frontLeftOf} and \pred{frontRightOf}}
\begin{align}
\pred{frontLeftOf}(S,O)\defiff{}&l_{SO}>2\nonumber\\
&\land r_{SO}>w(l_{SO}),\\
\pred{frontRightOf}(S,O)\defiff{}&l_{SO}>2\nonumber\\
&\land r_{SO}<-w(l_{SO}).
\end{align}

\paragraph{\pred{rearLeftOf} and \pred{rearRightOf}}
\begin{align}
\pred{rearLeftOf}(S,O)\defiff{}&l_{SO}<-2\nonumber\\
&\land r_{SO}>w(l_{SO}),\\
\pred{rearRightOf}(S,O)\defiff{}&l_{SO}<-2\nonumber\\
&\land r_{SO}<-w(l_{SO}).
\end{align}

\paragraph{\pred{overlapping}} With $\varepsilon_A=10^{-4}\,\mathrm{m}^2$,
\begin{equation}
    \pred{overlapping}(S,O)
    \defiff A_\cap(S,O)>\varepsilon_A.
\end{equation}

\paragraph{\pred{touching}} With geometric distance tolerance $\varepsilon_d=10^{-3}\,\mathrm{m}$,
\begin{equation}
\begin{aligned}
\pred{touching}(S,O)\defiff{}&A_\cap(S,O)\leq\varepsilon_A\\
&\land\Bigl(F_S\text{ touches }F_O\\
&\qquad\lor d_{fs}\leq\varepsilon_d\Bigr).
\end{aligned}
\end{equation}

\paragraph{\pred{veryNear} and \pred{near}}
\begin{align}
\pred{veryNear}(S,O)
&\defiff 0<d_{fs}(S,O)\leq2\,\mathrm{m},\\
\pred{near}(S,O)
&\defiff 2<d_{fs}(S,O)\leq5\,\mathrm{m}.
\end{align}
The four distance/contact states are mutually exclusive.

%%%%%%%%%%%%%%%%%%%%%%%%%%%%%%%%%%%%%%%%%%%%%%%%%%%%%%%%%%%%%%%%%%%%%%%%%%%%%%%%

\subsection{Motion Predicates}
\label{appsec:motion}

For an entity $E$, let $\mathbf{v}_E=[v_x,v_y]^\top$. Tracked-object velocities are interpreted in the common global frame. Ego velocity is converted from the ego-state representation to the global geometric-center velocity used by the predicate framework.

\paragraph{\pred{hasVelocity}, \pred{hasVelocityX}, \pred{hasVelocityY}, and \pred{hasSpeed}}
The planar velocity of entity $E$ is
\begin{equation}
    \mathbf{v}_E=[v_x,v_y]^\top .
\end{equation}
The predicate \pred{hasVelocity} stores this planar velocity, while \pred{hasVelocityX} and \pred{hasVelocityY} store its individual components. The corresponding speed magnitude is
\begin{equation}
    v_E=\sqrt{v_x^2+v_y^2}.
\end{equation}

\paragraph{\pred{hasAccelerationX}, \pred{hasAccelerationY}, and \pred{hasAcceleration}} For ego states exposing native planar acceleration,
\begin{equation}
    a_E=\sqrt{a_x^2+a_y^2}.
\end{equation}
The component predicates store $a_x$ and $a_y$.

\paragraph{\pred{hasRelativeSpeedTo}} The relative-speed magnitude is
\begin{equation}
    v^{rel}_{SO}=\|\Delta\mathbf{v}_{SO}\|_2.
\end{equation}

\paragraph{\pred{hasLongitudinalRelativeSpeedTo}} The subject-longitudinal relative velocity is
\begin{equation}
    v^{rel,long}_{SO}
    =\mathbf{e}_S^\top\Delta\mathbf{v}_{SO}.
\end{equation}

\paragraph{\pred{hasLateralRelativeSpeedTo}} The subject-lateral relative velocity is
\begin{equation}
    v^{rel,lat}_{SO}
    =\mathbf{n}_S^\top\Delta\mathbf{v}_{SO}.
\end{equation}

\paragraph{\pred{hasClosingSpeedTo}} The radial closing speed is
\begin{equation}
v^{close}_{SO}
=
-\frac{\Delta\mathbf{p}_{SO}^{\top}\Delta\mathbf{v}_{SO}}
       {\|\Delta\mathbf{p}_{SO}\|_2},
\qquad d_c(S,O)>0.
\end{equation}
Positive values mean decreasing center separation.

\paragraph{\pred{hasVelocityTowardTarget}} The component of subject velocity toward the current object center is
\begin{equation}
v^{\rightarrow O}_{S}
=
\frac{\mathbf{v}_S^\top\Delta\mathbf{p}_{SO}}
     {\|\Delta\mathbf{p}_{SO}\|_2},
\qquad d_c(S,O)>0.
\end{equation}

\paragraph{\pred{hasSubjectForwardSpeed}} The subject forward speed in its body-heading frame is
\begin{equation}
    v_S^{fwd}=\mathbf{e}_S^\top\mathbf{v}_S.
\end{equation}

\paragraph{\pred{hasVelocityHeading}} When $v_E\geq0.75\,\mathrm{m/s}$, the velocity heading is
\begin{equation}
    \theta_E^v=
    \operatorname{wrap}\!\left(\operatorname{atan2}(v_y,v_x)\right).
\end{equation}
Below this speed the velocity direction is considered insufficiently stable and the predicate is not emitted.

\paragraph{\pred{hasEffectiveTravelHeading}} This predicate stores a conservative travel direction selected from map heading, velocity heading, and displacement heading. A displacement heading is considered informative only after at least $0.40\,\mathrm{m}$ displacement; a velocity heading requires at least $0.75\,\mathrm{m/s}$. If both motion cues are available, they must agree within $0.45\,\mathrm{rad}$. An accepted motion-derived direction must agree with an available unambiguous map heading within $0.60\,\mathrm{rad}$. The selected angle is denoted $\theta_E^{travel}$.

\paragraph{\pred{hasTravelDirectionSource}} Stores the categorical provenance of $\theta_E^{travel}$, i.e., which accepted map/motion evidence supplied the selected direction.

\paragraph{\pred{hasTravelDirectionDifferenceTo}} When both participants have effective travel headings,
\begin{equation}
\Delta\theta^{travel}_{SO}
=
\left|
\operatorname{wrap}
(\theta_O^{travel}-\theta_S^{travel})
\right|.
\end{equation}

%%%%%%%%%%%%%%%%%%%%%%%%%%%%%%%%%%%%%%%%%%%%%%%%%%%%%%%%%%%%%%%%%%%%%%%%%%%%%%%%

\subsection{Temporal Predicates}
\label{appsec:temporal}

The continuity limit is
\begin{equation}
    T_{gap}=0.5\times1.5=0.75\,\mathrm{s}.
\end{equation}
Temporal derivatives and continuous-streak quantities are not bridged across larger gaps.

\paragraph{\pred{precedes}} For two instances of the same persistent track,
\begin{equation}
\begin{aligned}
\pred{precedes}(E_{k-1},E_k)\defiff{}&
\operatorname{sameTrack}(E_{k-1},E_k)\\
&\land\,0<\Delta t_k\leq0.75\,\mathrm{s}.
\end{aligned}
\end{equation}

\paragraph{\pred{hasDeltaTimeFromPrevious}} Stores $\Delta t_k$ in seconds for a valid preceding observation.

\paragraph{\pred{hasDisplacementFromPrevious}}
\begin{equation}
    d_k^{disp}=\|\mathbf{p}_k-\mathbf{p}_{k-1}\|_2.
\end{equation}

\paragraph{\pred{hasHeadingChangeFromPrevious}}
\begin{equation}
    \Delta\theta_k
    =\operatorname{wrap}(\theta_k-\theta_{k-1}).
\end{equation}

\paragraph{\pred{hasSpeedChangeFromPrevious}}
\begin{equation}
    \Delta v_k=v_k-v_{k-1}.
\end{equation}

\paragraph{\pred{hasEstimatedAcceleration}} For $\Delta t_k>0$,
\begin{equation}
    \hat a_k=\frac{\Delta v_k}{\Delta t_k}.
\end{equation}

\paragraph{\pred{hasDisplacementHeading}} When $d_k^{disp}\geq0.40\,\mathrm{m}$,
\begin{equation}
\theta_k^{disp}
=
\operatorname{atan2}(y_k-y_{k-1},x_k-x_{k-1}).
\end{equation}

\paragraph{\pred{hasObservedFrameCount} and \pred{hasObservedDuration}} These predicates store the count and elapsed duration of the entity history maintained by the current scenario processing state. The duration is measured from the first available observation in that maintained history to the current timestamp.

\paragraph{\pred{hasTotalObservedFrameCount} and \pred{hasTotalObservedSpan}} The total count includes all available entity observations in the current processing window, including observations separated by gaps. If $t_{first}$ is the first observation timestamp,
\begin{equation}
    T_E^{total}=t_k-t_{first}.
\end{equation}

\paragraph{\pred{hasContinuousObservedFrameCount} and \pred{hasContinuousObservedDuration}} The continuous count increments only while successive observations satisfy the $0.75\,\mathrm{s}$ continuity condition. When a gap exceeds that limit the streak restarts at one. If $t_{cont,0}$ is the start of the current streak,
\begin{equation}
    T_{E,k}^{cont}=t_k-t_{cont,0}.
\end{equation}

\paragraph{\pred{hasPairObservedFrameCount} and \pred{hasPairObservedDuration}} The same continuity logic is maintained for each directed pair. If $t_{pair,0}$ is the first timestamp of the current uninterrupted co-observation streak,
\begin{equation}
    T_{SO,k}^{pair}=t_k-t_{pair,0}.
\end{equation}

\paragraph{\pred{hasCenterDistanceChangeFromPrevious}}
\begin{equation}
    \Delta d_{c,k}=d_{c,k}-d_{c,k-1}.
\end{equation}

\paragraph*{\hspace*{-2em}m) \pred{hasFreeSpaceDistanceChangeFromPrevious}}
\begin{equation}
    \Delta d_{fs,k}=d_{fs,k}-d_{fs,k-1}.
\end{equation}

%%%%%%%%%%%%%%%%%%%%%%%%%%%%%%%%%%%%%%%%%%%%%%%%%%%%%%%%%%%%%%%%%%%%%%%%%%%%%%%%

\subsection{Map Predicates}
\label{appsec:map}

Let $M$ denote an HD-map polygon and let
\begin{equation}
\rho(E,M)
=
\frac{\operatorname{area}(F_E\cap M)}
     {\operatorname{area}(F_E)}
\end{equation}
be footprint overlap. Primary-map matching uses exact unbuffered map geometry. A candidate must contain the entity center or provide sufficient footprint overlap. The primary-map configuration uses a minimum overlap ratio $0.20$, an ambiguity margin $0.05$, and a lateral tie-break margin $0.50\,\mathrm{m}$.

\paragraph{\pred{inLane} and \pred{inLaneConnector}} Center-membership predicates are denoted below by $I_L$ and $I_C$, respectively:
\begin{align}
I_L(E,L)&\defiff\mathbf{p}_E\in\operatorname{polygon}(L),\\
I_C(E,C)&\defiff\mathbf{p}_E\in\operatorname{polygon}(C).
\end{align}

\paragraph{\pred{intersectsLane} and \pred{intersectsLaneConnector}} Writing $J_L$ and $J_C$ for the corresponding footprint-intersection predicates,
\begin{align}
J_L(E,L)&\defiff\operatorname{area}(F_E\cap L)>0,\\
J_C(E,C)&\defiff\operatorname{area}(F_E\cap C)>0.
\end{align}

\paragraph{\pred{hasPrimaryLane} and \pred{hasPrimaryLaneConnector}} These predicates store the unique conservative primary lane $L^*$ or lane connector $C^*$ selected from the exact candidate set. When competing candidates are too similar under the overlap/center/lateral/heading evidence, no forced primary assignment is made.

\paragraph{\pred{hasPrimaryMapOverlapRatio}} For the selected primary primitive $M^*$,
\begin{equation}
    \rho_E^*=\rho(E,M^*).
\end{equation}

\paragraph{\pred{hasAmbiguousMapMatch}} A positive Boolean indicator is emitted when candidate evidence is too ambiguous for a reliable primary map assignment under the $0.05$ ambiguity margin and associated deterministic tie-breaking logic.

\paragraph{\pred{hasBaselineProgress}} If $p_E^*$ is the nearest point on the selected directed baseline and $s(\cdot)$ its arc-length coordinate,
\begin{equation}
    s_E=s(p_E^*).
\end{equation}

\paragraph{\pred{hasBaselineLateralOffset}} For baseline pose $(x_b,y_b,\theta_b)$ at $p_E^*$,
\begin{equation}
r_E^{map}
=
-\sin\theta_b(x_E-x_b)
+\cos\theta_b(y_E-y_b).
\end{equation}
Positive values lie left of the directed baseline.

\paragraph{\pred{hasMapHeading}} Stores the local legal travel direction
\begin{equation}
    \theta_E^{map}=\operatorname{wrap}(\theta_b).
\end{equation}

\paragraph{\pred{hasBaselineCurvature}} Stores the local signed baseline curvature
\begin{equation}
    \kappa_E=\kappa(s_E).
\end{equation}

\paragraph{\pred{hasMapSpeedLimit}} Stores the speed limit of $M^*$ in $\mathrm{m/s}$ when the map primitive provides a finite valid value.

\paragraph{\pred{hasParentRoadblock} and \pred{hasParentRoadblockConnector}} These relations expose the parent roadblock of a selected lane and the parent roadblock connector of a selected lane connector, respectively.

\paragraph{\pred{inIntersection} and \pred{intersectsIntersection}} Writing $I_{int}$ and $J_{int}$ for center and footprint intersection,
\begin{align}
I_{int}(E,I)&\defiff\mathbf{p}_E\in\operatorname{polygon}(I),\\
J_{int}(E,I)&\defiff\operatorname{area}(F_E\cap I)>0.
\end{align}

\paragraph{\pred{hasPrimaryMapIntersection}} Stores the intersection topologically/geometrically associated with the selected primary lane or connector.

\paragraph{\pred{inCrosswalk} and \pred{intersectsCrosswalk}} Writing $I_{cw}$ and $J_{cw}$ for center and footprint crosswalk membership,
\begin{align}
I_{cw}(E,CW)&\defiff\mathbf{p}_E\in\operatorname{polygon}(CW),\\
J_{cw}(E,CW)&\defiff\operatorname{area}(F_E\cap CW)>0.
\end{align}

\paragraph{\pred{hasSpatialMapRelation}} For reliable primary map primitives $M_S^*$ and $M_O^*$, this categorical predicate stores whether they are the same primitive, left/right adjacent, successor/predecessor, or unrelated according to the directed map topology.

\paragraph{\pred{hasMapProgressDifferenceTo}} When both entities share the same selected baseline,
\begin{equation}
    \Delta s_{SO}=s_O-s_S.
\end{equation}

\paragraph{\pred{hasSignedPathDistanceTo}} On the same primary primitive,
\begin{equation}
    d_{SO}^{path}=s_O-s_S.
\end{equation}
If $O$ lies downstream on an accepted direct/connected successor path, the remaining arc length on the subject primitive and downstream progress are summed. Predecessor paths receive the corresponding negative sign. The search does not force a path through an ambiguous branch.

\paragraph{\pred{inSameLaneAs}}
\begin{equation}
\begin{aligned}
\pred{inSameLaneAs}(S,O)\defiff{}&
L_S^*=L_O^*\\
&\land\neg A_S^{map}\land\neg A_O^{map},
\end{aligned}
\end{equation}
where $A_E^{map}$ is the map-ambiguity flag. Travel-direction agreement is not required by this predicate.

\paragraph{\pred{sharesIntersectionWith}} Asserted when both entities have independently established geometric or topological association with the same intersection.

%%%%%%%%%%%%%%%%%%%%%%%%%%%%%%%%%%%%%%%%%%%%%%%%%%%%%%%%%%%%%%%%%%%%%%%%%%%%%%%%

%%%%%%%%%%%%%%%%%%%%%%%%%%%%%%%%%%%%%%%%%%%%%%%%%%%%%%%%%%%%%%%%%%%%%%%%%%%%%%%%

\subsection{Safety-Related Thresholds}
\label{appsec:safety_thresholds}

Safety-related thresholds for longitudinal following, collision risk, and lane-change gaps follow the criteria described in UN Regulation No.~157 (UN R157). For M1/N1 vehicles travelling at speed $v\leq60\,\mathrm{km/h}$, the minimum following distance is
\begin{equation}
    d_{\min}(v)=v\,t_{\mathrm{front}}(v),
\end{equation}
where $v$ is expressed in $\mathrm{m/s}$. The minimum time-gap values are
\begin{center}
\begin{tabular}{cccccccc}
\toprule
$v$ [km/h] & 7.2 & 10 & 20 & 30 & 40 & 50 & 60 \\
\midrule
$t_{\mathrm{front}}$ [s] & 1.0 & 1.1 & 1.2 & 1.3 & 1.4 & 1.5 & 1.6 \\
\bottomrule
\end{tabular}
\end{center}
with piecewise-linear interpolation between the specified values. For speeds below $2\,\mathrm{m/s}$, a minimum following distance of $2\,\mathrm{m}$ applies.

An imminent collision risk is identified when collision avoidance would require a braking demand of at least $5\,\mathrm{m/s^2}$. For regular lane changes, the approaching vehicle in the target lane shall not be required to decelerate by more than $3.0\,\mathrm{m/s^2}$ under the corresponding assessment conditions, and the inter-vehicle distance shall not become smaller than the distance travelled by the ALKS vehicle in $1.0\,\mathrm{s}$.

\subsection{Interaction Predicates}
\label{appsec:interaction}

Interaction predicates are deterministic temporal relations that combine geometric, kinematic, map, and history evidence. The auxiliary condition symbols below are abbreviations for the explicitly defined condition blocks.

\subsubsection{\pred{follows}}
For a participant $X$ projected onto a directed lane/connector path, let $I_X=[s_X^{rear},s_X^{front}]$ denote its footprint interval and
\begin{equation}
    g_{SO}=s_O^{rear}-s_S^{front}
\end{equation}
the forward bumper gap. The complete relation is
\begin{equation}
\begin{aligned}
\pred{follows}(S,O)\defiff{}&
\mathcal F_{\mathrm{veh}}
\land\mathcal F_{\mathrm{path}}
\land\mathcal F_{\mathrm{leader}}\\
&\land
(\mathcal F_{\mathrm{move}}\lor\mathcal F_{\mathrm{queue}})
\land\mathcal F_{\mathrm{persist}} .
\end{aligned}
\end{equation}
$\mathcal F_{\mathrm{veh}}$ requires compatible motor-vehicle participants. $\mathcal F_{\mathrm{path}}$ requires one reliable unambiguous directed lane/connector path, at most three topology hops, positive gap, non-overlapping footprints, and no reverse path motion beyond $0.30\,\mathrm{m/s}$. $\mathcal F_{\mathrm{leader}}$ requires $O$ to be the unique nearest valid forward leader; candidates within $0.50\,\mathrm{m}$ of one another are treated as ambiguous.

For moving traffic,
\begin{equation}
\begin{aligned}
\mathcal F_{\mathrm{move}}\defiff{}&
v_S^{path}>0.30\,\mathrm{m/s}\\
&\land 0<g_{SO}/v_S^{path}\leq5.0\,\mathrm{s}\\
&\land g_{SO}\leq80\,\mathrm{m},
\end{aligned}
\end{equation}
where $v_S^{path}$ is the subject velocity projected onto the directed path. For stop-and-go traffic,
\begin{equation}
\begin{aligned}
\mathcal F_{\mathrm{queue}}\defiff{}&
v_S^{path}\leq2.0\,\mathrm{m/s}\\
&\land v_O^{path}\leq4.0\,\mathrm{m/s}\\
&\land 0<g_{SO}\leq12\,\mathrm{m}.
\end{aligned}
\end{equation}
Finally,
\begin{equation}
    \mathcal F_{\mathrm{persist}}
    \defiff T_{\mathrm{cond}}\geq1.0\,\mathrm{s}.
\end{equation}
Thus, queue following is included in \pred{follows}, while the queue-specific case is additionally represented by the separate \pred{queuesBehind} predicate.

\subsubsection{\pred{queuesBehind}}

The predicate \pred{queuesBehind} represents the queue-specific case of longitudinal following. It uses the same vehicle, path, leader, and persistence requirements as \pred{follows}, while requiring the stop-and-go condition:
\begin{equation}
\begin{aligned}
\pred{queuesBehind}(S,O)\defiff{}&
\mathcal F_{\mathrm{veh}}
\land\mathcal F_{\mathrm{path}}
\land\mathcal F_{\mathrm{leader}}\\
&\land\mathcal F_{\mathrm{queue}}
\land\mathcal F_{\mathrm{persist}} .
\end{aligned}
\end{equation}

\subsubsection{\pred{changesLane}}
Let $C_s$ and $C_t$ denote the stable decoded source and target logical lane corridors and let $L_t$ be the emitted target lane. The implemented event can be summarized as
\begin{equation}
\begin{aligned}
\pred{changesLane}(S,L_t)\defiff{}&
\mathcal L_{\mathrm{hist}}
\land\mathcal L_{\mathrm{trans}}
\land\mathcal L_{\mathrm{motion}}\\
&\land\mathcal L_{\mathrm{target}}
\land\mathcal L_{\mathrm{score}} .
\end{aligned}
\end{equation}
The history gate is
\begin{equation}
\mathcal L_{\mathrm{hist}}
\defiff
N_{\mathrm{hist}}\geq1
\land T_{\mathrm{hist}}\geq0.5\,\mathrm{s}.
\end{equation}
The transition must be lateral rather than an ordinary longitudinal lane--connector continuation:
\begin{equation}
\begin{aligned}
\mathcal L_{\mathrm{trans}}\defiff{}&
C_s\neq C_t
\land\operatorname{Adjacent}(C_s,C_t)\\
&\land\neg\operatorname{Continuation}(C_s,C_t).
\end{aligned}
\end{equation}
The principal physical gate is
\begin{equation}
\begin{aligned}
\mathcal L_{\mathrm{motion}}\defiff{}&
v_S\geq0.10\,\mathrm{m/s}\\
&\land T_{\mathrm{trans}}\leq8.0\,\mathrm{s}\\
&\land|\Delta r|\geq1.0\,\mathrm{m}.
\end{aligned}
\end{equation}

The target normally requires two stable frames. Stable-lane evidence uses overlap ratio at least $0.45$. Physical onset uses lateral offset $0.15\,\mathrm{m}$, lateral velocity $0.15\,\mathrm{m/s}$, future target gain $0.40\,\mathrm{m}$, and two onset-support frames within a $6.0\,\mathrm{s}$ search window. Completion uses target overlap at least $0.60$, source remaining overlap at most $0.40$, and two stable completion frames; the enabled high-confidence single-frame completion requires target overlap at least $0.80$. A single-frame primary-target fallback requires overlap at least $0.70$.

Official map adjacency is preferred. The geometric fallback allows polygon separation at most $1.5\,\mathrm{m}$, centerline distance $2.0$--$6.5\,\mathrm{m}$, and direction difference at most $0.30\,\mathrm{rad}$; the local fallback uses centerline distance $1.25$--$6.5\,\mathrm{m}$ and direction difference at most $0.60\,\mathrm{rad}$. The decoder permits at most three unknown-gap frames and requires
\begin{equation}
\begin{aligned}
\mathcal L_{\mathrm{score}}\defiff{}&
(s_{\mathrm{LC}}\geq6.0
\land\neg C_{\mathrm{completion}})\\
&\lor
(s_{\mathrm{LC}}\geq3.25
\land E_{\mathrm{probable}}),
\end{aligned}
\end{equation}
where $s_{\mathrm{LC}}$ is the deterministic event score, $C_{\mathrm{completion}}$ indicates completion censoring, and $E_{\mathrm{probable}}$ denotes the enabled probable-event mode.

\subsubsection{\pred{mergesInFrontOf} and \pred{mergesBehind}}
Let $T$ denote the target stream of an already decoded lane change. The shared merge condition is
\begin{equation}
\begin{aligned}
\mathcal M_{\mathrm{base}}(S,O,T)\defiff{}&
\pred{changesLane}(S,T)
\land\mathcal M_{\mathrm{pre}}\\
&\land\mathcal M_{\mathrm{path}}
\land\mathcal M_{\mathrm{flow}}
\land\mathcal M_{\mathrm{stable}} .
\end{aligned}
\end{equation}
$\mathcal M_{\mathrm{pre}}$ requires $O$ to be present in the target stream before entry, using a $2.0\,\mathrm{s}$ pre-existence window and at least one supporting frame. $\mathcal M_{\mathrm{path}}$ requires target-path overlap at least $0.20$ and at most six topology hops. $\mathcal M_{\mathrm{flow}}$ requires travel-direction difference at most $0.55\,\mathrm{rad}$. $\mathcal M_{\mathrm{stable}}$ requires stable order for at least two frames within a $2.0\,\mathrm{s}$ confirmation window.

Let $O_{\mathrm{rear}}^*$ and $O_{\mathrm{front}}^*$ be the nearest valid target-stream neighbors behind and ahead of $S$. Then
\begin{equation}
\begin{aligned}
&\pred{mergesInFrontOf}(S,O)\\
&\quad\defiff
\mathcal M_{\mathrm{base}}(S,O,T)
\land S\succ_T O\\
&\qquad\land O=O_{\mathrm{rear}}^*
\land0<g_{SO}\leq40\,\mathrm{m}\\
&\qquad\land h_{\mathrm{rear}}\leq5.0\,\mathrm{s},
\end{aligned}
\end{equation}
and
\begin{equation}
\begin{aligned}
&\pred{mergesBehind}(S,O)\\
&\quad\defiff
\mathcal M_{\mathrm{base}}(S,O,T)
\land S\prec_T O\\
&\qquad\land O=O_{\mathrm{front}}^*
\land0<g_{SO}\leq40\,\mathrm{m}\\
&\qquad\land h_{\mathrm{front}}\leq5.0\,\mathrm{s}.
\end{aligned}
\end{equation}
Here $S\succ_T O$ and $S\prec_T O$ denote stable ahead/behind order on the target path, and $h$ is the corresponding post-merge headway when defined.

\subsubsection{\pred{crossesInFrontOf}}
Let $\mathbf d_S$ and $\mathbf d_O$ be unit vectors along the current headings. The two finite forward rays are
\begin{align}
R_S(\lambda)&=\mathbf p_S+\lambda\mathbf d_S,
&0\leq\lambda\leq10\,\mathrm{m},\\
R_O(\mu)&=\mathbf p_O+\mu\mathbf d_O,
&0\leq\mu\leq10\,\mathrm{m}.
\end{align}
For their unique finite-segment intersection $C$, the implemented relation is
\begin{equation}
\begin{aligned}
&\pred{crossesInFrontOf}(S,O)\\
&\quad\defiff
\mathcal X_{\mathrm{type}}
\land\mathcal X_{\mathrm{hist}}
\land\mathcal X_{\mathrm{local}}\\
&\qquad\land\mathcal X_{\mathrm{angle}}
\land\mathcal X_{\mathrm{ray}}
\land\mathcal X_{\mathrm{order}}\\
&\qquad\land\mathcal X_{\mathrm{map}} .
\end{aligned}
\end{equation}
$\mathcal X_{\mathrm{type}}$ requires distinct dynamic road users and rejects parked entities; pedestrian--pedestrian pairs are disabled by default. $\mathcal X_{\mathrm{hist}}$ requires at least three observations, simultaneous observation for at least $1.0\,\mathrm{s}$, and track displacement at least $0.30\,\mathrm{m}$. Locality requires $d_c(S,O)\leq30\,\mathrm{m}$, and
\begin{equation}
\mathcal X_{\mathrm{angle}}
\defiff
25^\circ\leq\Delta\theta_{SO}\leq155^\circ .
\end{equation}
The finite-ray condition is
\begin{equation}
\mathcal X_{\mathrm{ray}}
\defiff
C=R_S(\lambda_C)=R_O(\mu_C),
\quad
\lambda_C,\mu_C\in[0,10]\,\mathrm{m}.
\end{equation}
The subject speed must satisfy $v_S\geq0.30\,\mathrm{m/s}$. For a moving object,
\begin{equation}
t_S(C)=\lambda_C/v_S,\qquad
t_O(C)=\mu_C/v_O,
\end{equation}
and crossing order requires
\begin{equation}
\mathcal X_{\mathrm{order}}
\defiff
0.25\,\mathrm{s}
\leq t_O(C)-t_S(C)
\leq6.0\,\mathrm{s}.
\end{equation}
For an effectively stopped object, $t_O(C)$ is treated as unbounded after the parking filter.

The map/context filter uses a $12\,\mathrm{m}$ query radius, maximum road distance $6\,\mathrm{m}$, crosswalk-near distance $5\,\mathrm{m}$, and bicycle road-distance threshold $3\,\mathrm{m}$. Parked-object rejection uses speed at most $0.50\,\mathrm{m/s}$ for at least $2.0\,\mathrm{s}$, displacement at most $1.0\,\mathrm{m}$, and lane distance at most $1.5\,\mathrm{m}$. The conflict footprint/side buffer is $0.35\,\mathrm{m}$ and the side-search window is $3.0\,\mathrm{s}$.

\subsubsection{\pred{yieldsTo}}
The object must already be verified to cross in front of the subject. The complete rule is
\begin{equation}
\begin{aligned}
\pred{yieldsTo}(S,O)\defiff{}&
\pred{crossesInFrontOf}(O,S)
\land\mathcal Y_{\mathrm{type}}\\
&\land\mathcal Y_{\mathrm{compete}}
\land\mathcal Y_{\mathrm{concede}}\\
&\land\mathcal Y_{\mathrm{clear}}
\land\mathcal Y_{\mathrm{outside}}\\
&\land\mathcal Y_{\mathrm{proceed}}
\land\neg\mathcal Y_{\mathrm{red}} .
\end{aligned}
\end{equation}
The subject is vehicle-like and the object is a dynamic road user. The implementation uses a $4.0\,\mathrm{s}$ pre-event window and $4.0\,\mathrm{s}$ post-event verification window. Before concession, both users must approach the conflict with subject speed at least $0.50\,\mathrm{m/s}$, object speed at least $0.30\,\mathrm{m/s}$, and initial ETA difference
\begin{equation}
|\tau_S-\tau_O|\leq2.5\,\mathrm{s}.
\end{equation}
Let $v_b$ be the subject pre-response baseline speed. It must satisfy $v_b\geq1.0\,\mathrm{m/s}$ and the required speed drop is
\begin{equation}
\Delta v_{\mathrm{req}}
=
\max(0.75,\;0.15v_b)\quad[\mathrm{m/s}].
\end{equation}
The subject must remain within $18\,\mathrm{m}$ of the conflict and outside the conflict while $O$ clears it. Clearance uses half the object length plus a $0.25\,\mathrm{m}$ buffer. A relevant RED traffic signal invalidates the yield interpretation. After clearance,
\begin{equation}
\begin{aligned}
\mathcal Y_{\mathrm{proceed}}\defiff{}&
v_S^{post}\geq0.80\,\mathrm{m/s}\\
&\land D_S^{post}\geq1.50\,\mathrm{m}.
\end{aligned}
\end{equation}
The implementation also records a slow-state reference threshold of $1.25\,\mathrm{m/s}$ when evaluating the concession.

\subsubsection{\pred{overtakes}}
Overtaking is a complete, fully observed same-direction Case-1 maneuver:
\begin{equation}
\begin{aligned}
\pred{overtakes}(S,O)\defiff{}&
\mathcal O_{\mathrm{hist}}
\land\mathcal O_{\mathrm{start}}
\land\mathcal O_{\mathrm{flow}}\\
&\land\mathcal O_{\mathrm{depart}}
\land\mathcal O_{\mathrm{pass}}
\land\mathcal O_{\mathrm{reverse}}\\
&\land\mathcal O_{\mathrm{return}}
\land\mathcal O_{\mathrm{duration}} .
\end{aligned}
\end{equation}
$\mathcal O_{\mathrm{hist}}$ requires at least five common frames, at least $0.5\,\mathrm{s}$ pre-departure history, and frame gaps no larger than $1.5\,\mathrm{s}$. The pair search radius is $80\,\mathrm{m}$. $\mathcal O_{\mathrm{start}}$ requires the same logical corridor and initial subject-behind gap at least $1.0\,\mathrm{m}$.

Same-flow consistency requires
\begin{equation}
\begin{aligned}
\mathcal O_{\mathrm{flow}}\defiff{}&
\Delta\theta^{travel}_{SO}\leq0.55\,\mathrm{rad}\\
&\land f_{\mathrm{sameFlow}}\geq0.80 .
\end{aligned}
\end{equation}
During the passing phase,
\begin{equation}
\begin{aligned}
\mathcal O_{\mathrm{pass}}\defiff{}&
v_S\geq2.0\,\mathrm{m/s}\\
&\land v_O\geq0.50\,\mathrm{m/s}\\
&\land v_{SO}^{rel,long,+}\geq0.30\,\mathrm{m/s}\\
&\land |r_{SO}|_{\mathrm{side}}\geq1.25\,\mathrm{m}.
\end{aligned}
\end{equation}
The departure must use an official/validated adjacent passing corridor. Lane search uses radius $8.0\,\mathrm{m}$, minimum overlap ratio $0.10$, ambiguity margin $0.05$, and at most eight topology hops. Order reversal must produce at least $1.0\,\mathrm{m}$ full clearance, the subject must return stably to the original corridor for two frames, and
\begin{equation}
\mathcal O_{\mathrm{duration}}
\defiff T_{\mathrm{overtake}}\leq30\,\mathrm{s}.
\end{equation}
The approach, passing, and completion phases each require at least $0.5\,\mathrm{s}$.

\subsection{Traffic-Control Predicates}
\label{appsec:traffic}

\paragraph{\pred{controls}} Let $\mathcal M(s)$ denote the controlled lane-connector movements associated with traffic signal $s$. The relation is defined by
\begin{equation}
    \pred{controls}(s,m)\defiff m\in\mathcal M(s).
\end{equation}
The relation is independent of the current signal color.

\paragraph{\pred{hasSignalState}} At time $t$, the predicate stores the latest valid traffic-light state $q$ for signal $s$:
\begin{equation}
    \pred{hasSignalState}(s,q)\defiff q=Q_s(t).
\end{equation}

\paragraph{\pred{isRelevantSignal}} At time $t$, a traffic signal is relevant to agent $a$ only before entry into the controlled movement:
\begin{equation}
\begin{aligned}
&\pred{isRelevantSignal}(a,s)\\
&\quad\defiff\exists m:\pred{controls}(s,m)\\
&\qquad\land\operatorname{approaches}(a,m,t)\\
&\qquad\land\neg\operatorname{occupies}(a,m,t).
\end{aligned}
\end{equation}

The approach and occupancy terms refer to the implementation's current lane/connector path context; they do not introduce unreported numerical parameters. The predicate stops being emitted once the agent occupies the controlled connector.

%%%%%%%%%%%%%%%%%%%%%%%%%%%%%%%%%%%%%%%%%%%%%%%%%%%%%%%%%%%%%%%%%%%%%%%%%%%%%%%%

\subsection{Risk Predicates}
\label{appsec:risk}

\paragraph{\pred{hasConflictRiskWith}}

The emission condition can be summarized as
\begin{equation}
\begin{aligned}
&\pred{hasConflictRiskWith}(S,O)\defiff
\mathcal R_{\mathrm{dyn}}\land\mathcal R_{\mathrm{geom}}\\
&\qquad\land\mathcal R_{\mathrm{closing}}
\land\mathcal R_{\mathrm{interact}}
\land\mathcal R_{\mathrm{broad}}
\land\mathcal R_{\mathrm{future}}
\land\mathcal R_{\mathrm{safety}}.
\end{aligned}
\end{equation}
$\mathcal R_{\mathrm{dyn}}$ requires two dynamic road users. $\mathcal R_{\mathrm{geom}}$ requires valid oriented footprints and excludes already-overlapping pairs. $\mathcal R_{\mathrm{closing}}$ requires radial closing speed at least $0.50\,\mathrm{m/s}$. $\mathcal R_{\mathrm{interact}}$ is satisfied by same/connected lane-path topology, a VRU pair within $30\,\mathrm{m}$, a tight same-flow corridor ($|r^{travel}_{SO}|\leq2.5\,\mathrm{m}$, $d_c\leq45\,\mathrm{m}$, direction difference $\leq15^\circ$), or local cross/opposite motion ($d_c\leq40\,\mathrm{m}$ and direction difference $\geq25^\circ$).

Let $r_S$ and $r_O$ be footprint half-diagonals and $d_{\mathrm{CPA}}^{center}$ the minimum center distance under the same constant-velocity model. The broad-phase condition is
\begin{equation}
\mathcal R_{\mathrm{broad}}
\defiff
d_{\mathrm{CPA}}^{center}
\leq r_S+r_O+0.50\,\mathrm{m}.
\end{equation}
If $\tau^*$ and $d_{\mathrm{CPA}}^{fs}$ are the time and oriented-footprint clearance at the predicted minimum, then
\begin{equation}
\begin{aligned}
\mathcal R_{\mathrm{future}}\defiff{}&
0<\tau^*\leq H\\
&\land d_{\mathrm{CPA}}^{fs}\leq0.50\,\mathrm{m}\\
&\land d_{fs}(S,O)-d_{\mathrm{CPA}}^{fs}\geq1.0\,\mathrm{m},
\end{aligned}
\end{equation}
where $H$ is the configured prediction horizon.

The condition $\mathcal R_{\mathrm{safety}}$ requires the predicted conflict to satisfy the applicable safety criteria described in Sec.~\ref{appsec:safety_thresholds}. An imminent collision condition is identified when the required collision-avoidance braking demand is at least $5\,\mathrm{m/s^2}$. For lane-change-related conflicts, $\mathcal R_{\mathrm{safety}}$ additionally considers whether the target-lane vehicle would be required to decelerate by more than $3.0\,\mathrm{m/s^2}$ or whether the inter-vehicle gap would fall below the distance travelled by the ALKS vehicle in $1.0\,\mathrm{s}$.

The prediction assumes constant velocity and fixed heading. The base predicate implementation uses a $2.0\,\mathrm{s}$ horizon; the retained standard launcher passes $H=3.0\,\mathrm{s}$. The risk computation uses a $0.50\,\mathrm{m}$ predicted-clearance threshold, minimum radial closing speed $0.50\,\mathrm{m/s}$, minimum clearance reduction $1.0\,\mathrm{m}$, and 24 bounded optimization iterations. The relation is emitted once per unordered pair and frame and does not use observed future ground truth.

\section*{Acknowledgment}

This work was supported by the Deutsche Forschungsgemeinschaft (DFG, German Research Foundation) under Project FIP 135/1, Project Number 549102058.

\bibliographystyle{IEEEtran}
\bibliography{IEEEexample}

\end{document}